\documentclass[runningheads]{llncs}

\usepackage{eccv}

\usepackage{eccvabbrv}

\usepackage{graphicx}
\usepackage{booktabs}

\usepackage[accsupp]{axessibility}  

\usepackage{hyperref}

\usepackage{orcidlink}

\usepackage{arydshln}
\usepackage{tabularray}

\usepackage{ulem} 
\usepackage{tikz}
\usepackage{bbding}
\usepackage{cancel}

\newcommand{\notcheckmark}{{$\surd$}\textsuperscript{\textcolor{black}{\kern-0.35em{\bf--}}}}

\usepackage{array,multirow,graphicx}

\usepackage{amsmath,amsfonts,bm}
\usepackage{booktabs}

\definecolor{myblue2}{RGB}{150,195,250}

\usepackage{enumitem}
\usepackage{wrapfig}

\usepackage{colortbl}

\newcolumntype{x}[1]{>{\centering\arraybackslash}p{#1}}
\newcolumntype{y}[1]{>{\raggedright\arraybackslash}p{#1}}
\newcolumntype{z}[1]{>{\raggedleft\arraybackslash}p{#1}}
\newcommand{\tablestyle}[2]{\setlength{\tabcolsep}{#1}\renewcommand{\arraystretch}{#2}\centering\footnotesize}

\def\eqref#1{equation~\ref{#1}}

\def\1{\bm{1}}

\usepackage{amssymb}
\usepackage{pifont}
\newcommand{\cmark}{\ding{51}}%
\newcommand{\xmark}{\ding{55}}%

\DeclareMathAlphabet{\mathsfit}{\encodingdefault}{\sfdefault}{m}{sl}
\SetMathAlphabet{\mathsfit}{bold}{\encodingdefault}{\sfdefault}{bx}{n}

\begin{document}

\title{OpenIns3D: Snap and Lookup for \\ 3D Open-vocabulary Instance Segmentation} 

\titlerunning{OpenIns3D: 3D Open-vocabulary Instance Segmentation}

\author{Zhening Huang\inst{1}\orcidlink{0000-0003-0039-4970} \quad
Xiaoyang Wu\inst{2}\orcidlink{0009-0002-2277-7104} \quad
Xi Chen\inst{2}\orcidlink{0009-0008-5008-4720} \quad \\
Hengshuang Zhao\inst{2}\thanks{Corresponding author.}\orcidlink{0000-0001-8277-2706} \quad
Lei Zhu\inst{3,4}\orcidlink{0000-0003-3871-663X} \quad
Joan Lasenby\inst{1}\orcidlink{0000-0002-0571-0218}
}

\authorrunning{Z.Huang et al.}

\institute{$^1$University of Cambridge \quad
$^2$The University of Hong Kong
\\
$^3$Hong Kong University of Science and Technology 
\\
$^4$The Hong Kong University of Science and Technology (Guangzhou)}
\maketitle

 \begin{figure*}[h]
 \centering
 \vspace{-10mm}

\textcolor{magenta}{\texttt{https://zheninghuang.github.io/OpenIns3D/}}

 \includegraphics[width=1.0\textwidth]{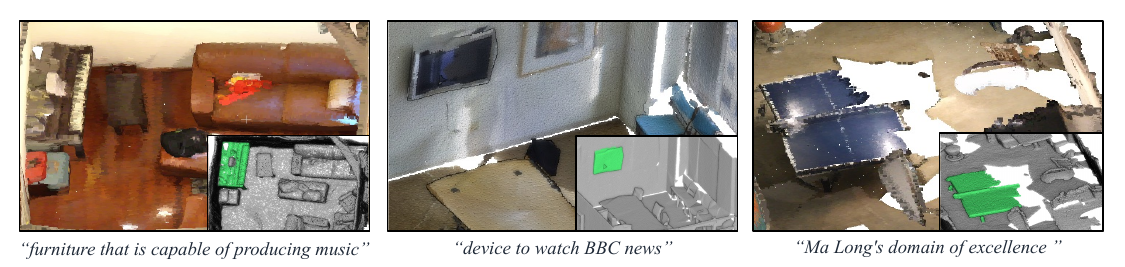}
 \centering
  \vspace{-6mm}
\caption{\textbf{Complex Queries 3D Instance Segmentation with OpenIns3D.} }

 \label{fig: intro_image}
\end{figure*}

\vspace{-8mm}
\begin{abstract}
In this work, we introduce OpenIns3D, a new 3D-input-only framework for 3D open-vocabulary scene understanding.
The OpenIns3D framework employs a “Mask-Snap-Lookup'' scheme. The “Mask'' module learns class-agnostic mask proposals in 3D point clouds, the “Snap'' module generates synthetic scene-level images at multiple scales and leverages 2D vision-language models to extract interesting objects, and the “Lookup'' module searches through the outcomes of “Snap'' to assign category names to the proposed masks.
This approach yet simple, achieves state-of-the-art performance across a wide range of 3D open-vocabulary tasks, including recognition, object detection, and instance segmentation, on both indoor and outdoor datasets.
Moreover, OpenIns3D facilitates effortless switching between different 2D detectors without requiring retraining. When integrated with powerful 2D open-world models, it achieves excellent results in scene understanding tasks. Furthermore, when combined with LLM-powered 2D models, OpenIns3D exhibits an impressive capability to comprehend and process highly complex text queries that demand intricate reasoning and real-world knowledge. 

\keywords{Open-Vocabulary Understanding \and 3D Scene Understanding \and Vision-Language Model}
\vspace{-8mm}
\end{abstract}

\vspace{-2mm}

 \begin{figure*}[ht]
 \centering
 \includegraphics[width=1.0\textwidth]{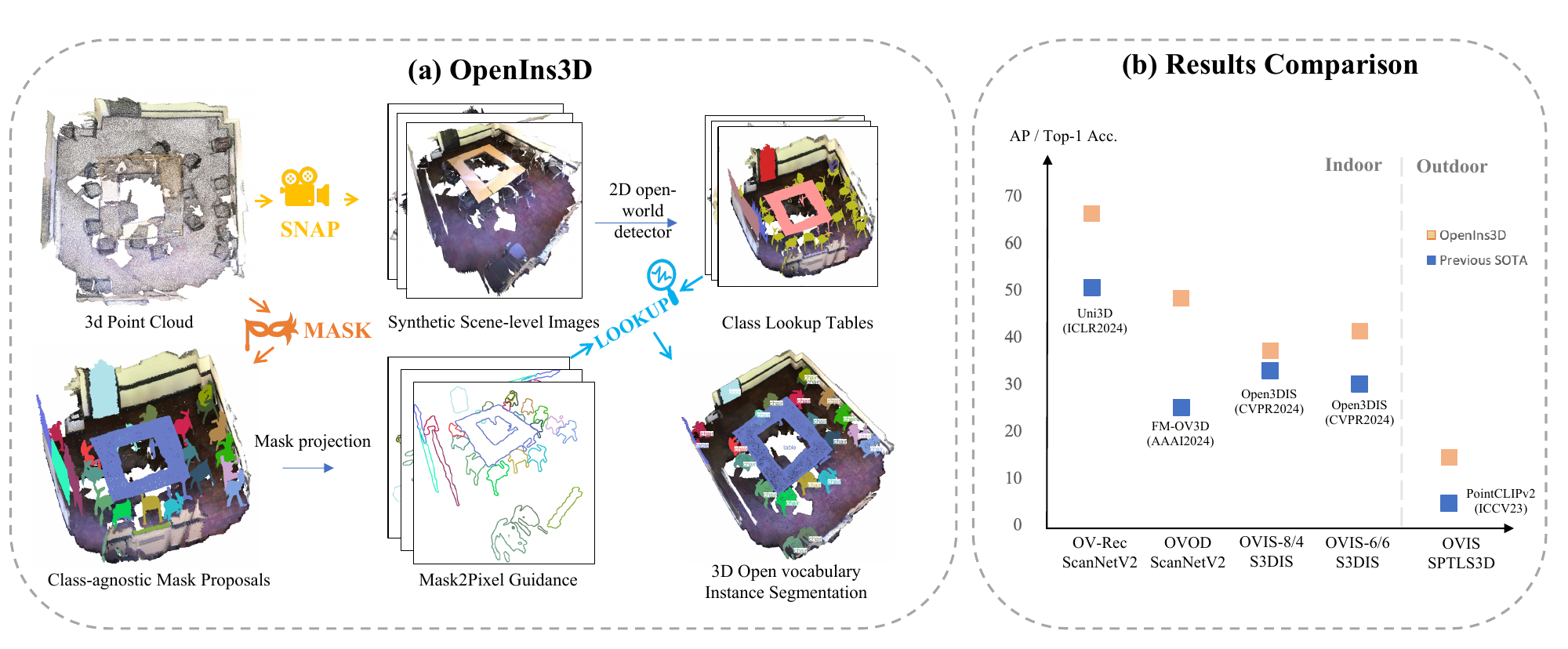}
 \centering
\vspace{-6mm}
  \caption{\textbf{High-level Illustrations of OpenIns3D and Quantitative Results}. (a) OpenIns3D follows the \textit{“Mask-Snap-Lookup”} steps for open-vocabulary scene understanding. (b) A list of SOTA results has been achieved on both indoor and outdoor datasets. OV-Rec: open-vocabulary object recognition. OVOD: open-vocabulary object detection. OVIS: open-vocabulary instance segmentation. PointCLIPV2 \cite{zhu2022pointclip}; Uni3D \cite{zhou2023uni3d}; Open3DIS \cite{nguyen2023open3dis}; FM-OV3D \cite{zhang2023fmov3d}}.
 \label{fig: msl-result}
 \vspace{-7mm}
\end{figure*}

\section{Introduction}

3D scene understanding plays a critical role in various domains, such as autonomous driving, robotic sensing, AR/VR, and manufacturing, among others. While the development of 3D closed-set understanding is relatively mature, scene understanding in an open-vocabulary setting is still in its infancy. Closed-set understanding can only handle a predefined set of concepts and scenarios but fails to provide valid responses when faced with unfamiliar concepts or variations in language usage. This limitation impacts its performance in dynamic and ever-changing contexts.

Thanks to internet-scale image-text datasets, significant progress has been made in 2D image open-vocabulary understanding \cite{radford2021learning, zhou2022maskclip, kirillov2023segment, chen2023openvocabulary, xu2023openvocabulary, brown2020gpt3, MaskCLIP, denseclipccl}. However, unlike 2D data that can be easily collected from the internet, constructing a large-scale 3D-text dataset poses a challenge. As a result, the most viable approach to achieving 3D open-vocabulary understanding involves leveraging 2D images to bridge language and 3D data. In this direction, there have been several notable works, such as OpenScene \cite{peng2022openscene}, PLA-family \cite{ding2022language, yang2023regionplc, ding2023lowis3d}, and CLIP2Scene \cite{chen2023clip2scene}. These works leverage well-aligned 2D images and 3D point clouds to conduct feature distillation or employ 2D caption models to construct 3D-text pairs. One prerequisite of these methods, however, is the availability of well-aligned 2D images and 3D point clouds. This means that posed 2D images, associated depth maps and camera models need to be accessible as inputs to the network. In real-life scenarios, there are numerous cases where 2D images or the information required to align 2D and 3D data are unavailable. For instance, to save storage space, point clouds generated with LiDAR are often stored without accompanying 2D images (example datasets include \cite{hackel2017isprs, tan2020toronto3d, roynard2017parislille3d}). In cases where point clouds are obtained from the registration of multiple scans from different sensors or are converted from 3D simulations/CAD models \cite{Mo_2019_CVPR, griffiths2019synthcity}, 2D images are often non-existent.

We believe that developing a 3D open-vocabulary framework without relying on well-aligned 2D images is meaningful, as this will simplify deployment prerequisites and enhance its applicability across a wide range of scenarios. To this end, we introduce \textit{OpenIns3D}, a framework designed to effectively perform 3D open-vocabulary scene understanding tasks without relying on 2D aligned images. Overall, OpenIns3D comprises three core steps: \textit{Mask}, \textit{Snap}, and \textit{Lookup}. An overall illustration of OpenIns3D is presented in Figure \ref{fig: msl-result}a.

\noindent\textbf{Mask}: Given a 3D point cloud, the first part of OpenIns3D learns class-agnostic mask proposals with a \textit{Mask Proposal Module} (MPM). This process is trained without any classification labels. To control the quality of the mask, MPM proposes a learnable \textit{Mask Scoring} module to predict the quality of each mask output and implements a list of \textit{Mask Filtering} techniques to discard invalid, low-quality masks. MPM outputs a list of class-agnostic masks in the scene.

\noindent\textbf{Snap}: Multiple synthetic scene-level images are generated with calibrated and optimized camera poses and intrinsic parameters. These images are specifically designed to encompass part or all of the relevant masks, aiming to minimize the need for multiple renderings. Instead of individually predicting the category of each mask proposal \cite{zhu2022pointclip,zhang2022pointclip, lu2023open}, the scene-level images are input into 2D open-vocabulary models for the simultaneous understanding of all interesting objects present in the scene. A Class Lookup Table (CLT) is then constructed to store all the detected object categories alongside their respective pixel locations.

\noindent\textbf{Lookup}: 
To precisely determine the positions of mask proposals in each image, Mask2Pixel maps are constructed. These maps project all 3D mask proposals onto 2D images with identical camera parameters used in \textit{Snap}. In the \textit{Lookup} phase, OpenIns3D searches through the CLT with the help of Mask2Pixel maps to precisely assign category names to 3D mask proposals. Results from multiple views are combined to establish initial mask classification outcomes. For remaining masks, a similar \textit{Lookup} procedure is carried out on a local scale to facilitate classification. Lastly, the 3D mask proposals are refined by removing masks lacking class assignments after both global and local \textit{Lookup}. 

OpenIns3D demonstrates promising performance in extensive comparisons with other methods, as summarized in Figure \ref{fig: msl-result}b. This simple and flexible approach achieves a range of state-of-the-art (SOTA) results on various 3D open-vocabulary tasks. Specifically, SOTA results are achieved on open-vocabulary instance segmentation for the indoor S3DIS \cite{armeni2016s3dis} dataset and the outdoor STPLS3D \cite{Chen_2022_BMVC} dataset. OpenIns3D even outperforms OpenMask3D \cite{takmaz2023openmask3d}, a concurrent work that heavily utilizes 2D images, for instance, segmentation on the challenging Replica \cite{straub2019replica} dataset. We also find that the “Snap and Lookup'' serves as a powerful open-vocabulary object recognition engine, which achieved SOTA object recognition tasks on ScanNet \cite{dai2017scannet}, outperforming the previous SOTA, a 1-billion parameter 3D foundation model \cite{zhou2023uni3d}. Lastly, when converting mask proposals into 3D bounding boxes, OpenIns3D also achieved state-of-the-art results in open-vocabulary object detection (OVOD) on ScanNet, outperforming previous image-dependent methods \cite{zhang2023fmov3d}. The design of OpenIns3D also allows 2D detectors to be changed without the need for retraining. This provides the model with the capability to evolve alongside the latest development of the 2D open-vocabulary models. Moreover, when 2D detectors are coupled with Large Language Models (LLMs), OpenIns3D can enable complex query understanding capability. When integrated with LISA \cite{reason_seg}, an LLM-powered reasoning segmentation model, OpenIns3D exhibits a strong ability to comprehend highly intricate language queries and performs reasoning segmentation in 3D, as illustrated in Figure \ref{fig: intro_image}. In summary, our contributions are:
\begin{itemize}[leftmargin=5mm, itemsep=0mm, topsep=-1mm, partopsep=-1mm]

\item OpenIns3D employs a distinct pipeline that operates without the need for well-aligned images. This approach achieves state-of-the-art results across a range of benchmarks and possesses the ability to comprehend highly complex input queries.

\item The proposed “Snap and Lookup” combination can serve as a powerful 3D object recognition engine, especially for noisy 3D objects extracted from scene-level scans. This capability is demonstrated by achieving state-of-the-art results with a large margin.
\end{itemize}

\section{Related work}
\noindent \textbf{3D open-vocabulary understanding.}
\label{3d-open-world}
\begin{figure}[t!]
 \centering
 \includegraphics[width=1.0\textwidth]{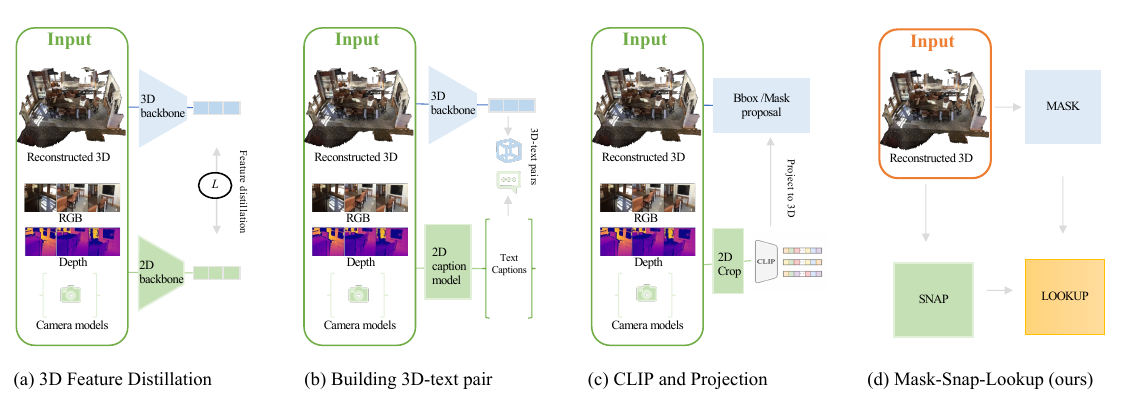}
  \vspace{-7mm}

\caption{\textbf{Four Categories of Open-Vocabulary 3D Scene Understanding Models.} a) 3D feature distillation frameworks, where 2D images are used as a bridge to distil language-aligned features into 3D, with typical works including OpenScene \cite{peng2022openscene} and Clip2Scene \cite{chen2023clip2scene}.
b) Building 3D-text pairs, where 2D captioning models are used to build 3D-text pairs for feature learning, with typical works including the PLA-family \cite{ding2023lowis3d, yang2023regionplc, ding2022language}
c) CLIP and Projection, where objects are cropped out of 2D images before being processed by CLIP, and the results are directly projected into 3D, including OpenMask3D \cite{takmaz2023openmask3d},  OV-3DET \cite{lu2023open}, CLIP$^2$\cite{zeng2023clip2} and Open3DIS \cite{nguyen2023open3dis}. d) OpenIns3D}

 \label{fig:2d-input-free}
  \vspace{-5mm}
\end{figure}
 Progress in 3D open-vocabulary understanding has been relatively slow compared to that of images. In the domain of 3D object classification tasks, methods like PointCLIP \cite{zhang2022pointclip}, PointCLIPV2 \cite{zhu2022pointclip}, and CLIP2Point \cite{huang2023clip2point} project 3D point clouds into depth maps and link them with 2D models for classification. However, these methods lack performance in scene-level understanding, where points are often overlapped and incomplete. For scene-level understanding, most work has primarily focused on leveraging well-aligned 2D posed images, depth maps, and point clouds \cite{chen2023clip2scene, peng2022openscene, rozenberszki2022language, ding2022language, yang2023regionplc, ding2023lowis3d, zeng2023clip2}. One notable example is OpenScene \cite{peng2022openscene}, which takes posed 2D images, depth maps, and 3D data as input, and feature distillation is performed to transfer 2D language-aligned features from images to 3D point clouds. Similarly, Clip2Scene \cite{chen2023clip2scene} builds dense pixel-point pairs by calibrating the LiDAR point cloud with corresponding images captured by six cameras. However, achieving instance-level understanding is challenging with these methods as they focus solely on semantic-level understanding. In contrast, PLA \cite{ding2022language} and its follow-up work RegionPLC \cite{yang2023regionplc} and Lowis3D \cite{ding2023lowis3d} utilize a 2D caption model to construct 3D-text pairs to learn features. However, the PLA-family works rely on a binary head to classify the input object into base categories or novel categories, and the transferability of this binary head to different base-novel splits is very limited, posing a challenge for flexible applications. One current work, OpenMask3D \cite{takmaz2023openmask3d}, utilizes well-aligned 2D images to learn features for mask proposals, leading to impressive results in open-vocabulary instance segmentation. Open3DIS \cite{nguyen2023open3dis} follows the same procedure, enhancing both mask proposals and mask classification with images, yielding better performance. A common issue with these methods is their reliance on well-aligned 3D and 2D pairs in the input, which may not always be available in real applications. We summarise the key difference between these methods in Figure \ref{fig:2d-input-free}. Simplifying input requirements improves method flexibility and compatibility. In this work, we explore how to conduct 3D open-vocabulary understanding without relying on 2D images.

\noindent \textbf{Image generation from 3D.}
Projection-based methods have been extensively explored in the past for 3D understanding and have proven to be beneficial for obtaining complementary features. For instance, MVCNN \cite{su15mvcnn} projects 3D objects to different views to aid in feature learning, while LAR \cite{bakr2022look} introduces object centre projection methods to generate images for 3D objects from various angles, assisting visual grounding tasks. Additionally, Virtual View Fusion \cite{kundu2020virtual} employs the original camera pose but enlarges the field of view, resulting in enhanced 2D feature transfer. However, these methods encounter challenges like best view selection, object occlusion, information loss during projection, and long rendering times. In the context of open-vocabulary settings, the quality of the projected image plays a crucial role in model performance. In our work, we evaluate different projection methods, along with their compatibility with 2D open-vocabulary models, to identify an optimal solution that achieves good results and is efficient to implement.

\section{OpenIns3D}

\subsection{Baseline and Challenges}

\noindent\textbf{Baseline.} We build a naive baseline by adopting the recent 3D instance segmentation backbone Mask3D \cite{schult2023mask3d} to generate mask proposals. To make the Mask Proposal Module (MPM) fit for the open-vocabulary setting, we remove all components in Mask3D that use the classification labels. Later, PointCLIP \cite{zhu2022pointclip} is adopted for mask understanding. This naive approach, although satisfying the requirement of 3D inputs only, has long rendering times and unsatisfactory performance (See in Table \ref{rendering_abl}.) There are several problems in this baseline model. 

\noindent\textbf{Challenge 1: excessive mask proposals.} in Mask3D \cite{schult2023mask3d}, mask proposals are filtered by the mask classification logit, which is removed in the class-agnostic setting. Therefore, an effective mask filtering scheme is needed. 

\noindent\textbf{Challenge 2: low quality of 3D instances.} 3D instances extracted from scene scans are typically broken, uncompleted, and sparse. Therefore, generated images by simple projection are not easily understood by 2D VL models. 

\noindent\textbf{Challenge 3: lack of context information.} Humans recognize imperfect 3D point cloud objects through scene comprehension. However, isolated instance point cloud projections lack this context. A solution is to project both the mask and background onto images, yet this introduces distracting elements, potentially confusing classification-level models like CLIP.

\noindent\textbf{Challenge 4: domain gap between projected images and natural images.} Rendered images often differ significantly from natural images used in training, posing a challenge for 2D visual language models to comprehend.

\subsection{Overall Framework}
 \begin{figure*}[t!]
 \vspace{-5mm}
 \centering
 \includegraphics[width=1.0\textwidth]{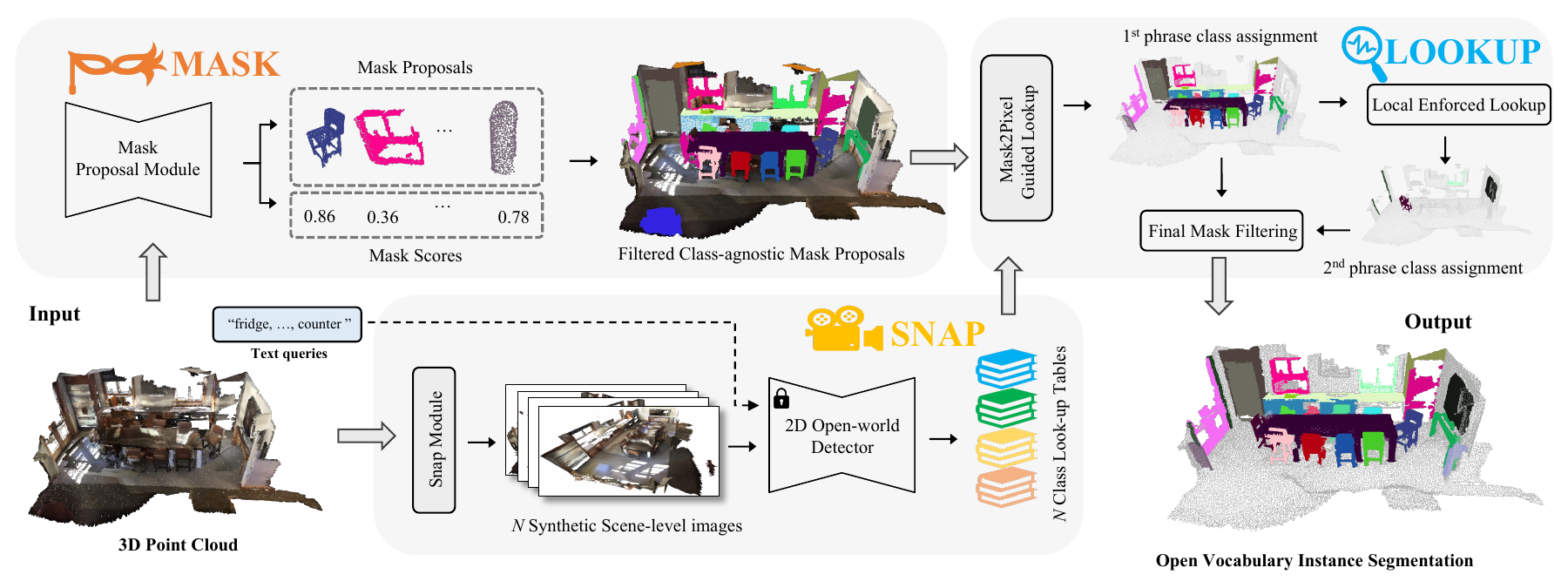}
 \centering
  \caption{\textbf{General Pipeline of OpenIns3D} OpenIns3D first processes point clouds with MPM to generate 3D mask proposals and mask scores. The \textit{Snap} module (detailed in Figure \ref{fig: snap}) then renders $N$ synthetic scene-level images, which are later passed into the 2D open-world model along with the input text queries. The detection results from the 2D model are stored in the \textit{Class Lookup Table} (CLT). Finally, both the mask proposals and CLT are fed into the \textit{Lookup} module, where \textit{Mask2Pixel Guided Lookup} (detailed in Figure \ref{fig: mask2point}) is performed at the global level, followed by a \textit{Local Enforced Lookup} at the local level to unlock the semantic meaning of mask proposals. The final mask filtering refines the mask proposals and obtains the final results.}
 \label{fig: architecture}
 \vspace{-4mm}
\end{figure*}

In this section, we present our design of OpenIns3D, which targets the aforementioned four problems. The pipeline of OpenIns3D is shown in Figure \ref{fig: architecture}. 
\subsubsection{Mask: class-agnostic mask proposal} \hfill \break

\noindent Within the Mask Proposal Module (MPM), we introduce two straightforward designs aimed at filtering low-quality masks generated in the baseline model.

\noindent\textbf{Mask scoring.}
Inspired by \cite{maskscoring2019huang, kirillov2023segment, chen2023openvocabulary}, We feed the instance queries generated from the Mask Module into a shallow MLP module to predict the quality, i.e IoU of the mask. The predicted IoU ($IoU_m$) is supervised by the ground truth IoU ($IoU_{gt}$) value during the training stages, which is calculated between the predicted mask and its matched ground truth mask in the Bipartite Matching. For unmatched prediction masks ($IoU_u$), we label the ground truth IoU value as zero. The loss is computed using $L2$. To avoid overly low IoU predictions, a hyper-parameter $\gamma$ is introduced to reduce the weight of loss for unmatched masks. Therefore, the total loss function for the MPM is:
\begin{equation}\label{eq:ms_loss}
\mathcal{L}_{\text{total}} = \gamma \sum ({IoU}_{u})^2 + \sum ({IoU}_m - {IoU}_{\text{gt}})^2
\end{equation}

\noindent\textbf{Mask filtering.}
To enhance mask quality, three filters are applied. Firstly, we retain masks with a model-predicted IoU score above a threshold of $\beta$, ensuring that only high-quality masks are kept. Secondly, drawing inspiration from SAM \cite{kirillov2023segment}, we focus on stable masks by comparing two binary masks derived from the same underlying soft mask using different threshold values. Specifically, we introduce an offset value $\alpha$ and select masks where the IoU between the pair of thresholded masks (one with $-\alpha$ and the other with $+\alpha$) exceeds 80\%. Lastly, small objects in the scene often lead to invalid proposals, so we filter out mask proposals that have a point number lower than \(N_{\text{min}}\). By employing these techniques, the quantity of mask proposals will decrease, resulting in cleaner and higher-quality masks for subsequent mask understanding tasks (\textit{Challenge 1}).

\subsubsection{Snap: Synthetic Scene-level View Generation}  \hfill \break

\noindent Rendering images from points can be a time-consuming task, especially when the number of rendering jobs is high. We propose a multi-scale synthetic scene-level image scheme.

\begin{figure*}[t!]
 \centering
\includegraphics[width=1.0\textwidth]{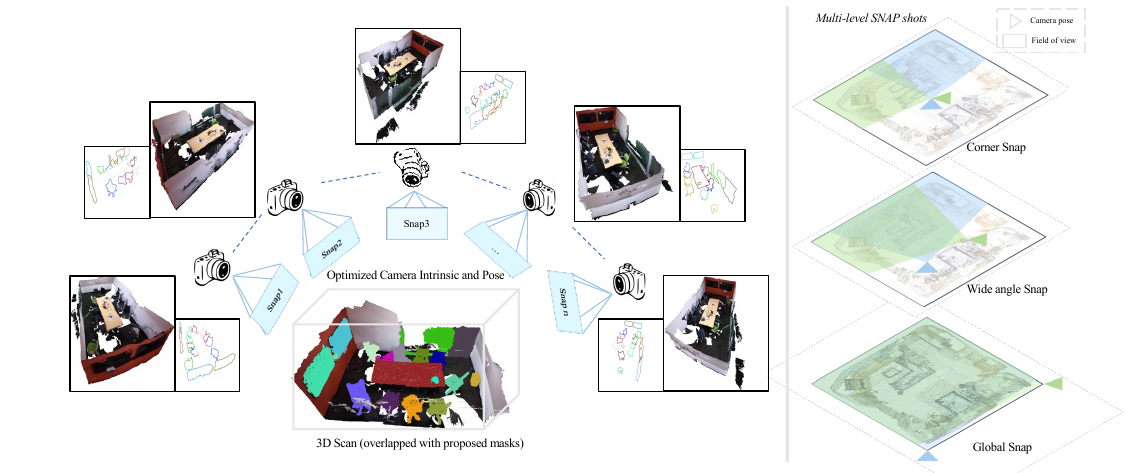}
 \centering
  \caption{\textbf{\textit{Snap} and Mask2Pixel Maps.} Multiscale snaps are conducted to render images with different levels of detail for scene understanding, including wide-corner snaps, wide-angle snaps, and global snaps. Cameras are positioned on the top of the scene and point towards the centre or corners, and the field of view is determined with the calibrated intrinsic matrix. With the defined camera models, Mask2Pixel maps are built to store the location of each 3D mask in the 2D image (using the same colour to represent 2D-3D correspondences) to guide the search for category names.}
 \label{fig: snap}
 \vspace{-4mm}
\end{figure*}

\noindent\textbf{Camera pose selection.} The Snap module captures scene-level images at three scales: global, corner, and wide-angle, as shown in Figure \ref{fig: snap}. For global-level images, cameras are positioned above the scene and point directly toward the centre of the scene. For corner images, cameras are positioned above the centre and point toward the corner, while for the wide-angle images, cameras are positioned at the 3 by 3 grid interaction points, pointing toward the furthest corner. Using the camera position coordinates \(P_{\text{cam}}\), target coordinates \(P_{\text{target}}\), and the up axis of the scene \(U\), the \textit{Lookat} function can be employed to determine the pose matrix \(Pose\). A more detailed mathematical formulation of this is presented in the supplementary materials.

\noindent\textbf{Camera intrinsic calibration.} Once the camera extrinsic matrix is established, the fields of view for the captured images are adjusted by modifying the camera's intrinsic parameters. The goal is to ensure that the entire scene or a specific part of it is encompassed within the captured images. To achieve this goal, we initialize an arbitrary camera intrinsic matrix and then adjust the focal lengths ($f_x$ and $f_y$) and the principal point coordinates ($c_x$ and $c_y$) through scaling. The scaling is performed by readjusting the projected areas in the image coordinate space. For example: if the projected points of the pre-defined area were located in image coordinates within the range of [$-1000$, $-192$] in the \(x\)-domain, our calibrated intrinsic parameters transform this range to [0, 1000] in \(x\). Importantly, we preserve the aspect ratio between the \(x\) and \(y\) coordinates to maintain the proportions of the final image without any distortion. This procedure ensures that each captured image is fully utilized and encompasses all regions of interest within the scene (\textit{Challenge 2 \& 3}).

\noindent\textbf{Class lookup table.} Upon obtaining \(N\) synthetic scene-level images, we input them into a 2D open-vocabulary detector. With text queries provided for interested classes, a list of detected objects in synthetic images can be obtained. Subsequently, information about detected objects, including their location and class, are stored in a designated \textit {Class Lookup Table} (CLT). This table will later be retrieved to allocate class categories to 3D mask proposals (\textit{Challenge 4}).

\subsubsection{Lookup: Mask Classification through Searching} 

\hfill \break

\noindent We conduct a multi-level search to assign category labels to the mask proposals generated from \textit{Mask} step.

\begin{wrapfigure}{bri}{0.5\textwidth}  
  \vspace{-12mm}
  \begin{center}
    \includegraphics[width=0.5\textwidth]{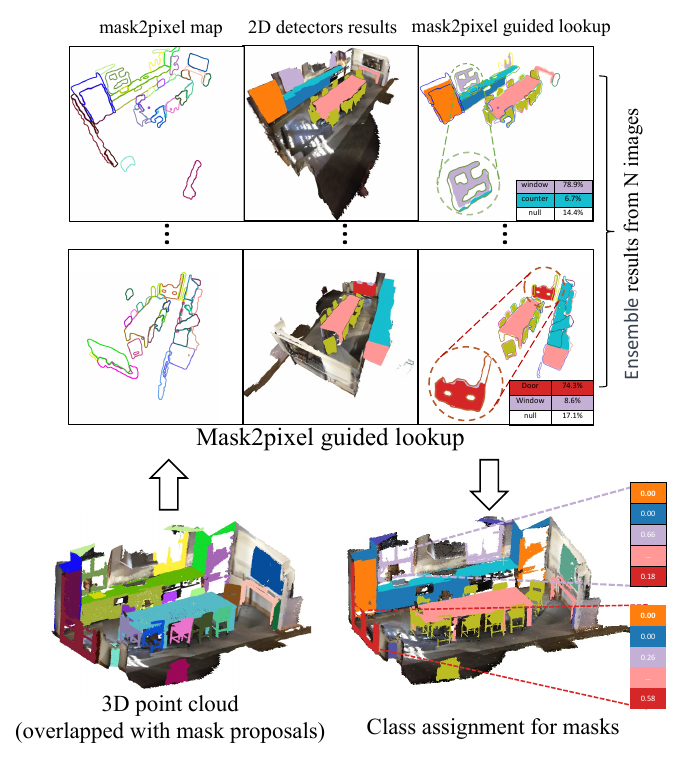}
  \end{center}
    \vspace{-4mm}
    \caption{\textbf{Mask2Pixel Guided Lookup Illustration.} IoUs between the 2D detection results and the projected masks are the guidance to assign class names to 3D masks. Multiple image results are ensembled.}
  \label{fig: mask2point}
  \vspace{-4mm}
\end{wrapfigure}

\noindent\textbf{Mask2Pixel guided lookup.} We introduce a \textit{Mask2Pixel Guided Lookup} (MGL) to search within CLT. The concept involves projecting each 3D mask proposal onto a 2D plane using the same camera extrinsic and intrinsic matrices that are utilized to generate the 2D image, as depicted in Figure \ref{fig: snap}. With knowledge of the precise pixel locations of each mask in images, we can conduct an accurate search through the CLT to identify the most likely class for each mask. The development of MPM takes into account occlusion by integrating depth information. To accomplish the matching, we follow a three-step approach:
1.	based on the mask's projection onto the 2D plane, we select the best-matched class categories in terms of IoU values; 
2.	if the IoU value of the best-matched object on the 2D plane is below 20\%, the match is disregarded.
3.	we aggregate results from multiple views to formulate the final prediction, calculating probability scores using their normalized average IoU values, as illustrated in Figure \ref{fig: mask2point}.

\noindent\textbf {Local enforced lookup.}
While the \textit{Mask2Pixel Guided Lookup} assigns class categories to mask proposals, some masks may not correspond to objects in the CLT. To address this, we introduce a \textit{Local Enforced Lookup} (LEL) approach. We crop out the remaining masks from 2D scene-level images using enlarged bounding boxes and process them with the 2D detector to encourage detection. To select the best views, we introduce an \textit{Occlusion Report} method to assess occlusion conditions for each mask in each projection, and then choose the top $K$ views for LEL. More details for \textit{Occlusion Report} can be found in the supplementary materials.

\noindent\textbf{Final mask refinement.}
With the previous lookup approaches, a large proportion of mask proposals obtain a category prediction. All masks that have no category predictions after the MGL and LEL stages are eliminated.

\vspace{-6mm}
\section{Experiments}
\noindent\textbf{Datasets and evaluation scheme.}
We tested OpenIns3D on five datasets, including four indoor datasets, namely S3DIS \cite{armeni_cvpr16}, ScanNetv2 \cite{dai2017scannet}, ScanNet200\cite{rozenberszki2022language}, Repica \cite{straub2019replica}, and one outdoor dataset, STPLS3D \cite{Chen_2022_BMVC}. Among them, S3DIS, ScanNetv2, and ScanNet200 are indoor point cloud datasets generated from RGB-D images, Replica is a photo-realistic 3D indoor scene reconstruction, while STPLS3D is an aerial photogrammetry-constructed outdoor dataset. \textit{We exclusively used the 3D data with colour from these datasets and did not employ any 2D images, poses, or depth maps.} Following the settings of prior work \cite{ding2022language}, we excluded the “other furniture'' class in ScanNetv2 and the “clutter'' class in S3DIS due to their vague meanings. Replica and ScanNet200 are evaluated by following the settings of OpenMask3D \cite{takmaz2023openmask3d}. For STPLS3D, we merged the low, medium, and high vegetation classes into one “vegetation'' class and kept all the rest.

\begin{table}[t]
\begin{center}
    \caption{\textbf{Zero-shot object classification on ScanNetv2.} OpenIns3D's Snap and Lookup approach for mask classification, surpasses all previous methods, including the latest language-aligned large-scale 3D foundation model \cite{zhou2023uni3d}.}
\resizebox{\textwidth}{!}{

\begin{centering}
\label{tab:scannet-reco}
\begin{tabular}{c|c|ccccccccccccccccc}
\toprule
Method & Avg.  & Bed & Cab & Chair & Sofa & Tabl & Door & Wind & Bksf & Pic & Cntr & Desk& Curt & Fridg & Bath & Showr & Toil & Sink  \tabularnewline
\midrule
PointCLIP \cite{zhu2022pointclip} & 6.3 & 0.0  & 0.0 & 0.0 &0.0  & 0.7 &0.0  & 0.0 & 91.8 & 0.0 &0.0 &0.0  &15.0  &0.0  &0.0 &0.0 &0.0 &0.0  \tabularnewline
PointCLIP V2 \cite{zhang2022pointclip} & 11.0 & 0.0  & 0.0 & 23.8 &0.0  & 0.0 &0.0  & 0.0 & 7.8 & 0.0 &\textbf{90.7} &0.0  &0.0  &0.0  &0.0 &64.4 &0.0 &0.0  \tabularnewline
CLIP2Point \cite{huang2023clip2point} & 24.9  & 20.8 & 0.0 & 85.1 & 43.3 & 26.5 &  69.9 & 0.0& 20.9 & 1.7 & 31.7 & 27.0 & 0.0 & 1.6 & 46.5&0.0&22.4&25.6    \tabularnewline
PointCLIP w/  TP. &26.1&0.0&\textbf{55.7}&72.8&5.0&5.1&1.7&0.0&\textbf{77.2}&0.0&0.0&51.7&0.3&0.0&0.0&40.3&85.3&49.2\tabularnewline
CLIP2Point w/  TP. & 35.2  &11.8&3.0&45.1&27.6&10.5&\textbf{61.5}&2.6&71.9&0.3&33.6&29.9&4.7& 11.5&\textbf{72.2}&92.4&86.1&34.0\tabularnewline
CLIP${^2}$ \cite{zeng2023clip2} & 38.5 & 32.6& {67.2}& 69.3 & 42.3 & 18.3 & 19.1 & 4.0 & 62.6 & 1.4 & 12.7 & 52.8 & 40.1 & 9.1 & 59.7 & 41.0 & 71.0 & 45.5  \tabularnewline
{Uni3D \cite{zhou2023uni3d}} & {45.8} & {58.5}& 3.7& {78.8} & {\textbf{83.7}} & {\textbf{54.9}} & 31.3 & {39.4} & 70.1 & {\textbf{35.1}} & 1.9 & {27.3} & {\textbf{94.2}} & 13.8 & 38.7 & 10.7 & 88.1 & {47.6}  \tabularnewline
\midrule
\rowcolor[gray]{.9}
\textbf{OpenIns3D} & \textbf{60.8} & \textbf{85.2} & 27.4 & \textbf{87.6} & 77.3 & 46.9 & 54.8 & \textbf{64.2 }& 71.4 & 9.9 & 80.8 &\textbf{ 82.7 }& 71.6 &\textbf{ 61.4 }& 38.7 & 0.0 & \textbf{87.9} & \textbf{85.7 }
\tabularnewline
\bottomrule
\end{tabular}
\end{centering}}
  \vspace{-5mm}
	\end{center}
\end{table}

\noindent\textbf{Implementation details.} For the S3DIS, ScanNetv2, Scannet200, and STPLS datasets, the MPM module is trained without utilizing any category labels, and $\lambda$ is set to 0.1 to reduce the weight of zero-IoU. The mask proposal of Replica is trained on ScanNet200, followed by OpenMask3D \cite{takmaz2023openmask3d}. The Snap module captures images with a size of 
\(1000 \times 1000\) including 16 global snaps, 4 corner snaps, and 4 wide-angle snaps. The top 0.5 m of the scene is removed for S3DIS, as the rooms are enclosed. For STPLIS3D, we followed Mask3D to split the large outdoor scene into patches of \(50m \times 50m\) and lifted the camera up to 10m. More implementation details are presented in the supplementary materials.

\begin{table}[t]
  \tablestyle{2.7pt}{1.0}
\caption{\textbf{3D Open-vocabulary Instance Segmentation Results on S3DIS and ScanNetv2.} We compare our zero-shot performance on the novel categories defined in the PLA-family work. Significant improvements are achieved on the S3DIS dataset, and competitive results are observed on ScanNetv2 (B/N: Base/Novel).}
\scriptsize
  \begin{tabular}{y{30mm} |x{10mm}x{10mm}x{10mm} |x{10mm}x{10mm}x{10mm}  |x{13mm}}\toprule
    OVIS &\multicolumn{3}{c}{S3DIS} & \multicolumn{3}{c}{ScanNetv2} &\\ \midrule
    Method & B/N &AP50 &AP25 & B/N &AP50 &AP25 & require 2D \\  \midrule
    PLA \cite{ding2022language} &8/4 &08.6 &- &10/7 &21.9 &- &\cmark \\
    RegionPLC  \cite{yang2023regionplc} &8/4 &- &- &10/7 &32.3 &- &\cmark \\
    Lowis3D \cite{ding2023lowis3d} &8/4 &13.8 &- &10/7 &31.2 &- &\cmark \\
    Open3DIS \cite{nguyen2023open3dis} &8/4 &26.3 &- &10/7 &- &- &\cmark \\
    
    Mask3d+PointClip \cite{zhu2022pointclip}  &--/4 &05.4 &10.3 &--/7 &04.5 &07.8 &\xmark \\ 
     \rowcolor[gray]{.9}  {OpenIns3D} &--/4 &\textbf{37.0} &\textbf{39.3}  &--/7 &{27.9} &\textbf{42.6}  &\xmark \\  
    \rowcolor[gray]{.9}  \tiny{\textcolor{teal}{\textit{improvement}}}  &--/4 & \tiny{\textcolor{teal}{(+10.7)}} & \tiny{\textcolor{teal}{(+29.0)}}  &--/7 & -- & \tiny{\textcolor{teal}{(+34.8)}}  &\xmark \\  \midrule
    PLA \cite{ding2022language} &6/6 &09.8 &- &8/9 &25.1 &- &\cmark \\
    RegionPLC \cite{yang2023regionplc}&6/6 &- &- &8/9 &32.2 &- &\cmark \\
    Lowis3D  \cite{ding2023lowis3d}  &6/6 &15.8 &- &8/9 &38.1 &- &\cmark \\
    Open3DIS  \cite{nguyen2023open3dis}  &6/6 &29.0 &- &8/9 &- &- &\cmark \\
    Mask3d+PointClip \cite{zhu2022pointclip}   &--/6 &08.5 &10.6 &--/9 &05.6 &06.7 &\xmark \\
  \rowcolor[gray]{.9}    \rowcolor[gray]{.9} {OpenIns3D} &--/6 &\textbf{33.0}  &\textbf{38.9} &--/9 &{19.5} &\textbf{27.9}  &\xmark \\ 
   \rowcolor[gray]{.9} \tiny{\textcolor{teal}{\textit{improvement}}} &--/6 & \tiny{\textcolor{teal}{(+4.0)}}  &\tiny{\textcolor{teal}{(+28.3)}} &--/9 &-- &\tiny{\textcolor{teal}{(+21.2)}} &\xmark \\ \midrule

  Mask3d+PointClip \cite{zhu2022pointclip} &--/12 &08.6 &09.3 &--/17 &04.5 & 14.4 & \xmark \\

 \rowcolor[gray]{.9}  OpenIns3D &--/12 &\textbf{28.3} &\textbf{29.5}  &--/17 &\textbf{28.7}  &\textbf{38.9} &\xmark \\
\rowcolor[gray]{.9}  \tiny{\textcolor{teal}{\textit{improvement}}}  &--/12 & \tiny{\textcolor{teal}{(+19.7)}} & \tiny{\textcolor{teal}{(+20.2)}} &--/17 &\tiny{\textcolor{teal}{(+24.2)}} &\tiny{\textcolor{teal}{(+24.5)}} &\xmark \\
\midrule
  \end{tabular}
  \label{tab: indoor_ovis}

\captionsetup{aboveskip=2pt}\captionsetup{belowskip=0pt}\captionof{table}{Open-vocabulary Object Detection ($AP_{25}$) on unseen classes in ScanNet.}
\label{tab:scannet_ovod}

\tablestyle{1.0pt}{1.0}
\scriptsize
\begin{tabular}{l|c|cccccccccc}

\toprule
{Methods} & {mean} & {toilet} &{bed} &{chair} &{sofa} &{dresser} &{table} &{cabinet} &{bookshelf} &{pillow} &{sink} \\
\midrule
OV-PointCLIP~\cite{zhang2022pointclip}  &3.1  &6.6 &2.3 &6.3 &3.9 &0.7 &7.2 &0.7 &2.1 &0.6 &0.8\\
OV-Image2Point~\cite{xu2022image2point} &0.8  &0.2 &0.8 &1.0 &1.4 &0.2 &2.8 &1.0 &0.9 &0.0 &0.1 \\
Detic-ModelNet~\cite{Detic}  &1.7  &4.2 &1.0 &4.6 &1.2 &0.2 &3.2 &0.6 &1.2 &0.0 &0.7\\
Detic-ImageNet~\cite{Detic} \hspace{0.25cm}  &0.4  &0.0 &0.0 &0.2 &0.0 &0.5 &1.8 &0.5 &0.3 &0.0 &0.7\\
OV-3DETIC~\cite{lu2023open} \hspace{0.25cm} &12.7 &49.0 &2.6 &7.3 &18.6 &2.8 &14.3 &2.4 &4.5 &3.9 &21.1  \\

L3Det \cite{zhu2023object2scene}  &{24.6} &{56.3} & {36.2} &{16.1} &{23.0} &{8.1} &{23.1} &{14.7} &{17.3} &{23.4} &{27.9} \\
FM-OV3D \cite{zhang2023fmov3d} &21.5 & 55.0 & 38.8 &19.2 &41.9 &23.8 & 3.5 &0.4 &6.0 & 17.4 &8.8  \\
\midrule
\rowcolor[gray]{.9}

OpenIns3D & \textbf{43.7}  & \textbf{79.5} & \textbf{70.5} & \textbf{76.9} & 15.8 & 0.0 & \textbf{53.1} & \textbf{40.1} & \textbf{41.2} & 7.1 &\textbf{ 53.1}  \\
\rowcolor[gray]{.9}
\tiny{\textcolor{teal}{\textit{improvement}}} & 
\tiny{\textcolor{teal}{{(+19.1)}}}&
\tiny{\textcolor{teal}{{(+23.2)}}}&
\tiny{\textcolor{teal}{{(+31.7)}}}&
\tiny{\textcolor{teal}{{(+57.7)}}}&
\tiny{\textcolor{teal}{{--}}}&
\tiny{\textcolor{teal}{{--}}}&
\tiny{\textcolor{teal}{{(+30)}}}&
\tiny{\textcolor{teal}{{(+25.4)}}}&
\tiny{\textcolor{teal}{{(+23.9)}}}&
\tiny{\textcolor{teal}{{}}}&
\tiny{\textcolor{teal}{{(+25.2)}}}\\

\bottomrule
\end{tabular}
\end{table}

\noindent\textbf{Open-vocabulary point cloud recognition.} We first evaluate OpenIns3D performance among all existing 3D open-vocabulary models with the same setting, i.e. \textbf{using only 3D inputs}. These models are most commonly tested on recognition tasks, including PointCLIP(v1\&v2) \cite{zhu2022pointclip} \cite{zhang2022pointclip}, Clip2Point\cite{huang2023clip2point}, CLIP$^2$\cite{zeng2023clip2}. We also compare with large-scale 3D foundation models, such as Uni3D \cite{zhou2023uni3d}, which has 1-billion parameters, and was trained with large-scale 3D shapes and image-text pairs. We followed their evaluation scheme and reported the Top-1 accuracy of instance classification on ScanNetv2. The results are presented in Table \ref{tab:scannet-reco}.

\begin{table}[!t]
    \begin{minipage}{0.55\textwidth}
    \centering
    \caption{\textbf{OVIS on the indoor Replica dataset.}}
    \tablestyle{0pt}{1.4}
    \scriptsize
    \begin{tabular}{y{38mm}|x{6mm}x{8mm}x{8mm}x{8mm}}
    \toprule
    Model & 2D & AP & AP$_{50}$  & AP$_{25}$ \\
    \midrule
    OpenScene \cite{peng2022openscene} (2D Fusion)& \cmark &$10.9$ & $15.6$ & $17.3$ \\ %
    OpenScene \cite{peng2022openscene} (2D/3D Ens.)& \cmark & $8.2$ & $10.4$& $13.3$ \\ %
    OpenMask3D \cite{takmaz2023openmask3d} & \cmark  & ${13.1}$ & ${18.4}$& ${24.2}$  \\ %
    \hline
    OpenScene \cite{peng2022openscene} (3D Distill) & \xmark  & $8.2$ & $10.5$& $12.6$ \\ %
  \rowcolor[gray]{.9}     OpenIns3D  & \xmark & \textbf{13.6} & \textbf{18.0} &\textbf{19.7}  \\
  \rowcolor[gray]{.9}
\tiny{\textcolor{teal}{\textit{improvement}}} & \xmark & \tiny{\textcolor{teal}{{(+5.4)}}}&
\tiny{\textcolor{teal}{{(+7.5)}}}&
\tiny{\textcolor{teal}{{(+7.1)}}}\\
    \bottomrule
    \end{tabular}
    \label{tab: Replica}
    \end{minipage}
    \hspace{0.7mm}
    \begin{minipage}{0.4\textwidth}
    \caption{OVIS on outdoor STPLS3D dataset}
    \centering
        \tablestyle{0pt}{1.96}
        \scriptsize
        \begin{tabular}{y{25mm}|x{8mm}x{8mm}x{8mm}}\toprule
        Model  & AP & AP$_{50}$  & AP$_{25}$ \\ \midrule
        PointCLIP \cite{zhu2022pointclip} & 02.0 & 02.6 & 04.0 \\ %
        PointCLIPv2 \cite{zhang2022pointclip} & 02.1 & 03.1&  05.2 \\ \hline
   \rowcolor[gray]{.9}        OpenIns3D & \textbf{11.4}& \textbf{14.2 }& \textbf{17.2}  \\ 
    \rowcolor[gray]{.9}        \tiny{\textcolor{teal}{\textit{improvement}}} & \tiny{\textcolor{teal}{{(+9.3)}}} & \tiny{\textcolor{teal}{{(+11.9)}}} & \tiny{\textcolor{teal}{{(+12.0)}}} \\ 
        \bottomrule
        \end{tabular}
            \label{tab: STPLS3D}
    \end{minipage}

\vspace{5mm}

\centering
\setlength{\tabcolsep}{5pt}
\caption{\textbf{3D instance segmentation results on the ScanNet200 validation set.} OpenIns3D demonstrates robust performance when compared to 2D-input-free models. However, notable limitations emerge when dealing with small objects in the common and tail classes.}
\tablestyle{4.0pt}{1.4}
\scriptsize
\begin{tabular}{l|c|ccc|ccc}
\toprule
Model & use 2D & $\text{AP}_{\text{head}}$ & $\text{AP}_{\text{common}}$ & $\text{AP}_{\text{tail}}$ &  AP & AP$_{50}$  & AP$_{25}$ \\
\midrule
OpenScene (2D Fusion) \cite{peng2022openscene}   & \cmark & $13.4$ &$11.6$& $9.9$ &$11.7$ & $15.2$& $17.8$ \\ %
OpenScene (2D/3D Ens.) \cite{peng2022openscene}  & \cmark & $11.0$ &$3.2$& $1.1$  & $5.3$ & $6.7$& $8.1$ \\ %
OpenMask3D  & \cmark & ${17.1}$ &${14.1}$ & ${14.9}$   & ${15.4}$ & ${19.9}$& ${23.1}$  \\  \hline
OpenScene (3D Distill)   & \xmark & $10.6$ & $2.6$ & $0.7$  & $4.8$ & $6.2$& $7.2$ \\ %
  \rowcolor[gray]{.9}  OpenIns3D &  \xmark & $\mathbf{16.0}$ &$\mathbf{6.5}$ & $\mathbf{4.2}$  & $\mathbf{8.8}$ & $\mathbf{10.3}$& $\mathbf{14.4}$   \\ %
\rowcolor[gray]{.9} \tiny{\textcolor{teal}{\textit{improvement}}} & \xmark & \tiny{\textcolor{teal}{{(+5.4)}}} & \tiny{\textcolor{teal}{{(+3.9)}}} & \tiny{\textcolor{teal}{{(+3.6)}}} & \tiny{\textcolor{teal}{{(+4.0)}}} & \tiny{\textcolor{teal}{{(+4.1)}}} & \tiny{\textcolor{teal}{{(+7.2)}}} \\

      \bottomrule
\end{tabular}

\label{tab:inst_seg_scannet200}
\vspace{-4mm}
\end{table}

\noindent\textbf{Open-vocabulary instance segmentation.}
We adopted various comparison schemes to align with existing methods. For 3D Open-vocabulary Instance Segmentation, we compared with PLA \cite{ding2022language}, and its follow-up works RegionPLC \cite{yang2023regionplc} and Lowis3D \cite{ding2023lowis3d}. For a fair comparison, we followed their category splits and compared our results on novel classes, as demonstrated in Table \ref{tab: indoor_ovis}. For STPLS3D, we compared OpenIns3D with baseline models whose classification module is PointCLIP and PointCLIPV2 \cite{zhu2022pointclip} (Table \ref{tab: STPLS3D}).
We also explored the performance of OpenIns3D on a more challenging dataset with more class categories. Specifically, we compared the performance of OpenIns3D with OpenMask3D, OpenScene, on Replica (Table \ref{tab: Replica}) as well as ScanNet200 (Table \ref{tab:inst_seg_scannet200}). Following OpenMask3D, we used Mask3D to generate mask proposals for OpenScene for evaluation.

\noindent\textbf{Open-vocabulary object detection.} Since there were limited works on 3D open-vocabulary instance segmentation at the time of conducting this work, we also selected some of the latest methods in the 3D open-world object detection domain for a more comprehensive evaluation. The evaluation is performed by converting generated masks into axis-aligned bounding boxes, and the results are shown in Table \ref{tab:scannet_ovod}.

\section{Results and Discussion}

\subsection{Comparison with SOTA}

First of all, OpenIns3D demonstrates impressive performance in open-vocabulary point cloud recognition, surpassing all previous methods, including the large-scale 3D foundation model by 15\%. This proves the effectiveness of the zero-shot “Snap'' and “lookup'' scheme. With the enhanced recognition capability, the performance of 3D open-vocabulary Object Detection among the ScanNet dataset has also achieved state-of-the-art results by a large margin.
For 3D instance segmentation, compared to works in the PLA family \cite{ding2023lowis3d, yang2023regionplc, ding2022language} and the latest work Open3DIS \cite{nguyen2023open3dis}, OpenIns3D does not require aligned images as input, (still need RGB information for points), achieves higher results on the S3DIS dataset, both in the 4 novel categories split and the 6 novel categories split. In SPTLS3D, OpenIns3D outperforms the baseline model PointCLIPV2 by 9.3 \% in AP. On the Replica dataset, OpenIns3D even outperformed OpenMask3D, which relies on well-aligned images for mask understanding. In the case of ScanNet200, OpenIns3D attains the highest performance compared to all other 3D input baselines. However, we have observed a decline in performance on tail and common classes within ScanNet200. The decrease in performance can be attributed to the low-quality and unclear reconstruction of smaller objects in the ScanNet200 scene. This is a noticeable limitation of OpenIns3D, but for better reconstructions, such as in Replica, and for Head classes within ScanNet200, OpenIns3D still demonstrates decent performance.

In summary, OpenIns3D demonstrates the best performance among all existing methods if only 3D data is used as input and outperforms many existing state-of-the-art methods that require 2D images. It also shows certain limitations on small objects that are not well-reconstructed in 3D scenes. 

\subsection{Ablation study}

\noindent\textbf{Mask quality ablation.} Following the evaluation on ScanNetv2, we assessed the class-agnostic mask quality using the average precision (AP) score. We treated all classes as universal since the predictions are class-agnostic. The evaluation was conducted on the ScanNetv2 validation set. Table \ref{tab: mask} demonstrates the effectiveness of the \textit{Mask Scoring} and \textit{Mask Filtering} designs.

\noindent\textbf{Multi-view ablation.}
We also studied the effects of using different numbers of views (Table \ref{tab: view_num}). Increasing the number of views used in the \textit{Lookup} module leads to better results. Additionally, Look Enforced Lookup provided a final improvement to the results.

\begin{table}[t]
\begin{center}
    
\caption{\textbf{Rendering and Inference Time Ablations.} Results tested on typical ScanNet scenes with 50 masks. OpenIns3D requires less rendering and inference time.}
\label{rendering_abl}
\tablestyle{3.3pt}{1.44}
\scriptsize
\begin{tabular}{l|ccc|ccc|c}\toprule
Rendering & Num of & 2D backbone & Img size & $T_{render}$ & $T_{infer}$ & $T_{total}$ & AP25 \\
& Img needed &  & (w $\times$ h) & (s/scene) & (s/scene)  & (s/scene) & (\%) \\ \midrule
PointCLIP & 250 & CLIP & $128^2$ & $5.2$ & $15.3$ & $20.5$ & $9.3$ \\
LAR & 250 & CLIP & $128^2$ & $14.3$ & $18.7$ & $33.0$ & $10.5$ \\
Mask rendering & 250 & CLIP & $128^2$ & $42.6$ & $19.5$ & $62.1$ & $7.3$ \\ \hline
\rowcolor[gray]{.9} OpenIns3D (ours) & 8 & G-DINO & $1000^2$ & $2.3$ & $6.2$ & $8.5$ & $29.8$ \\
\rowcolor[gray]{.9} OpenIns3D (ours) & 8 & ODISE & $1000^2$ & $2.3$ & $8.2$ & $10.5$ & $35.1$ \\
\bottomrule
\end{tabular}
\end{center}
    \begin{minipage}{0.55\textwidth}
    \centering
    \label{Replica}
    \caption{\textbf{MPM Ablation.} MS: Mask scoring. MF: Mask filtering.}
    \tablestyle{0pt}{1.2}
    \scriptsize
    \begin{tabular}{y{35mm}|y{15mm}y{15mm}}
        \toprule
        Method & AP50 & AP25 \\
        \midrule
        Mask3d-Supervised & 74.7 & 80.9 \\ \hline
        CA-Mask3d & $47.5$ & $49.2$ \\
        CA-Mask3d + MS & $50.2$ \tiny{\textcolor{teal}{$(+02.7)$}} & $53.3$ \tiny{\textcolor{teal}{$(+04.1)$}} \\
        CA-Mask3d + MF & $61.6$ \tiny{\textcolor{teal}{$(+14.1)$}} & $71.0$ \tiny{\textcolor{teal}{$(+21.8)$}} \\
      \rowcolor[gray]{.9}     CA-Mask3d + MS + MF & $64.6$ \tiny{\textcolor{teal}{$(+17.0)$}} & $73.4$ \tiny{\textcolor{teal}{$(+24.2)$}} \\
        \bottomrule
    \end{tabular}
    \label{tab: mask}
    \end{minipage}
    \hspace{1.5mm}
    \vspace{-3mm}
    \begin{minipage}{0.45\textwidth}
    \caption{\textbf{Number of Views Ablation.} LEL: Local Enforced Lookup}
    \centering
        \tablestyle{0pt}{1.44}
        \scriptsize
        \begin{tabular}{x{7mm}|x{6mm}x{6mm}x{6mm}x{8mm}x{9mm}x{9mm}}
        \toprule
        IDX & 4 & 8 & 16 & LEL & AP50 & AP25 \\
        \midrule
        i & \cmark & & & & $18.3$ & $27.1$ \\
        ii & & \cmark & & & $22.7$ & $35.1$ \\
        iii & & & \cmark & & $24.8$ & $37.5$ \\
        \rowcolor[gray]{.9} iv & & & \cmark & \cmark & $28.7$ & $38.9$ \\
        \bottomrule
    \end{tabular}
    \label{tab: view_num}
    
    \end{minipage}
\vspace{-5mm}

\end{table}

\noindent\textbf{Projection and 2D backbone ablation.} We conducted a comprehensive study on various rendering methods and their interaction with the 2D backbone to identify a suitable approach. We report the rendering time and inference performance for each method (Table \ref{rendering_abl}), and more details can be found in the supplementary materials. The key observation is that the scene-level rendering and understanding approach excels in speed while also demonstrating strong performance. Switching from Grounding Dino\cite{liu2023grounding} to the latest ODISE\cite{xu2023openvocabulary} also brings gains in performance, indicating that the OpenIns3D framework can easily benefit from the rapid development of 2D open-world detectors.

\noindent\textbf{Cross-domain analysis.} To evaluate the generalization capability of MPM across different domains, we trained and tested OpenIns3D on two different datasets, as shown in Table \ref{tab: cross_domain}. The cross-domain models also demonstrate impressive performance on both datasets when compared with the baseline. Notably, within the 17 classes of ScanNetv2, 11 classes do not exist in S3DIS. OpenIns3D, trained on S3DIS, still achieves decent performance among these unseen classes. 

\begin{figure*}[h]
 \centering
\vspace{-5mm}
\includegraphics[width=\textwidth]{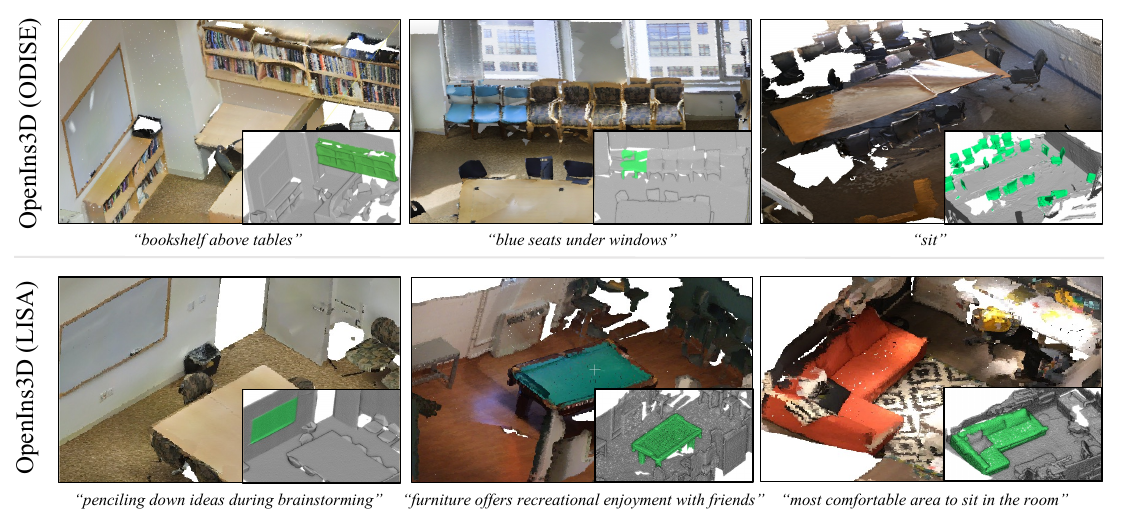}
 \centering
 \vspace{-5mm}
  \caption{
\textbf{Qualitative Results from OpenIns3D.} OpenIns3D (ODISE) demonstrates the ability to manage a versatile vocabulary. OpenIns3D (LISA) can conduct 3D reasoning segmentation.}
 \label{fig: qualitative}
\end{figure*}

\noindent\textbf{Free-flow language capability.} The snap and lookup scheme outsources the mask understanding tasks to a 2D vision-language model. Therefore when integrated with a 2D model powered by large language models (LLMs) like LISA \cite{reason_seg}, OpenIns3D can perform reasoning-based segmentation tasks (as depicted in Figure \ref{fig: qualitative}). For instance, when given the query "pencilling down ideas during brainstorming," OpenIns3D accurately segments the whiteboard, while for "furniture offers recreational enjoyment with friends," it precisely identifies and segments the pool table.

\vspace{-4mm}

\begin{table}[h]
\centering
\vspace{-10mm}
\caption{\textbf{Cross-domain Ablation.} We trained and tested OpenIns3D on two different datasets to examine its cross-domain capability. While S3DIS and ScanNetV2 have non-overlapping classes, OpenIns3D demonstrates decent generalization capability.}
\label{tab: cross_domain}
\tablestyle{5.0pt}{1.0}
\scriptsize
\begin{tabular}{x{20mm}|x{30mm}x{19mm}x{12mm}x{12mm}}\toprule
Test&Model &Training Data &AP50 &AP25 \\\midrule
\multirow{3}{*}{ScanNetv2 \cite{dai2017scannet}}
&Mask3D-Pointclip \cite{zhu2022pointclip} &ScanNetv2 &$04.5$ &$14.4$ \\
&OpenIns3D &ScanNetv2 &$28.7$ &$38.9$ \\
&OpenIns3D &S3DIS &\underline{$21.5$} &\underline{$33.6$} \\ \hline
\multirow{3}{*}{S3DIS \cite{armeni2016s3dis}}
&Mask3D-Pointclip \cite{zhu2022pointclip} &S3DIS &$03.5$ &$06.8$ \\
&OpenIns3D &S3DIS &$28.3$ &$29.5$ \\
&OpenIns3D &ScanNetv2  &\underline{$14.2$} &\underline{$19.8$} \\
\bottomrule
\end{tabular}
\vspace{-10mm}
\end{table}

\section{Conclusion}
\vspace{-3mm}

Achieving 3D open-vocabulary scene understanding is a challenging task, primarily due to the lack of extensive 3D-text data. Currently, most work in this domain focuses on using 2D images to bridge the gap between 3D and language. This, however, not only requires a good alignment between 2D and 3D but also evolves slowly due to the significant effort needed for retraining when changing the 2D backbone. OpenIns3D introduces a new pipeline, i.e. Mask-Snap-Lookup, for this task. The \textbf{Mask} module generates authentic masks in the 3D domain, while \textbf{Snap} renders scene-level images in 2D domains, and the \textbf{Lookup} module links the results from 2D to 3D precisely. This pipeline requires no image input, achieves better performance, and can evolve seamlessly with 2D models without training. We hope our work will provide a fresh perspective for researchers working towards open-world 3D scene understanding.

\section*{Acknowledgments}
This work is supported by the Girton College Graduate Research Awards at the University of Cambridge, School of Technology, National Highways sponsored through EPSRC Centre for Doctoral Training, the InnoHK funding launched by Innovation and Technology Commission, Hong Kong SAR, the National Natural Science Foundation of China No. 62201484, HKU Startup Fund, and HKU Seed Fund for Basic Research. We would like to express our gratitude to Yunhan Yang for his assistance in exploring rendering techniques and to Chengyao Wang for sharing his implementation of Mask3D.

\appendix
\section*{Appendix}
In a nutshell, OpenIns3D is a new pipeline for 3D open-world scene understanding that consumes only 3D coloured point clouds as input, making it easier to deploy in a wide range of scenarios. We also present detailed per-category results on how OpenIns3D performs on the ScanNetv2 \cite{dai2017scannet}, S3DIS \cite{armeni2016s3dis}, and STPLS3D \cite{Chen_2022_BMVC} datasets for both Instance Segmentation and Object Detection, for future work to compare with. In the development process, we tested a wide range of rendering methods and documented their different performances, which demonstrates how scene-level rendering stands out from other rendering methods. More detailed implementation details and methodologies of OpenIns3D are also presented. The section structure is listed as follows:

\begin{itemize}
    \item Section \ref{section1}: More Details on Methodologies
    \item Section \ref{section2}: Implementation Details
    \item Section \ref{section7}: Additional Experiments: OpenIns3D with RGBD
    \item Section \ref{section3}: Per-Categories Results
    \item Section \ref{section4}: Other Attempts for Image Rendering
    \item Section \ref{section5}: Limitation and future work
    \item Section \ref{section6}: More Visualization
\end{itemize}

\section{More Details on Methodologies}
\label{section1}

\noindent\textbf{Class-agnostic mask proposal module.}
We modified modules that require classification labels in Mask3D \cite{schult2023mask3d} to make it a class-agnostic setting. This includes 1. removing semantic probability components in Hungarian Matching, 2. eliminating semantic classification loss, 3. discarding classification logits-based ranking, and 4. getting rid of classification logits-based filtering. Instead, we added the \textit{Mask Scoring} and \textit{Mask Filtering} module to acquire high-quality mask proposals.

\noindent\textbf{Local enforced lookup.}
\label{occlusion report}
Here, we provide a detailed explanation of the \textit{Occlusion Report} module that we proposed to effectively evaluate the occlusion condition of masks in all synthetic images. Specifically, the following four steps are executed:
\begin{itemize}
 \item \textbf{Step 1. Point Count Array:} We initiate the process by constructing a 3D array with dimensions \(W \times H \times (M+1)\), where \(M\) represents the number of masks, and \(+1\) is for the background points. This array will be denoted as \(PC\), \textit{i.e.} point count, as it is designed to store the number of points of the 3D mask projected onto each pixel in the images. For example, if the pixel at coordinates \(i,j\) is occupied by two points from the 3D mask \(k\) during the projection, \(PT_{i,j,k}\) will be assigned the value 2.

 \item \textbf{Step 2. Foremost Point Identification:} Utilizing the depth map generated during the projection process, we construct a 2D array named \(FP\) with dimensions \(W \times H\), which is used to identify the foremost point in each pixel and indicate the originating mask number. For example, if pixel {i,j}'s foremost point is projected from Mask \(k\), we denote \(FP_{i,j} = k\).

 \item \textbf{Step 3. Occlusion Rate Calculation:} To evaluate the occlusion rate ($OR$) for mask $k$ within specific images, we compute the following formula:
 
 \[
 OR_{k} = \frac{\sum_{i=1}^{W} \sum_{j=1}^{H} PC_{i,j,k} \cdot (FP_{i,j} = k)}{T_{k}}
 \]

 where $T$ represent the total number of point in mask $k$.
 \item \textbf{Step 4. All Images Report: } Finally, we repeat steps 1-3 for all images to obtain an overall report of the occlusion rate of each mask across all images, forming the final \textit{Occlusion Report}.
\end{itemize}

\begin{figure*}[t!]
 \centering
 \includegraphics[width=1.0\textwidth]{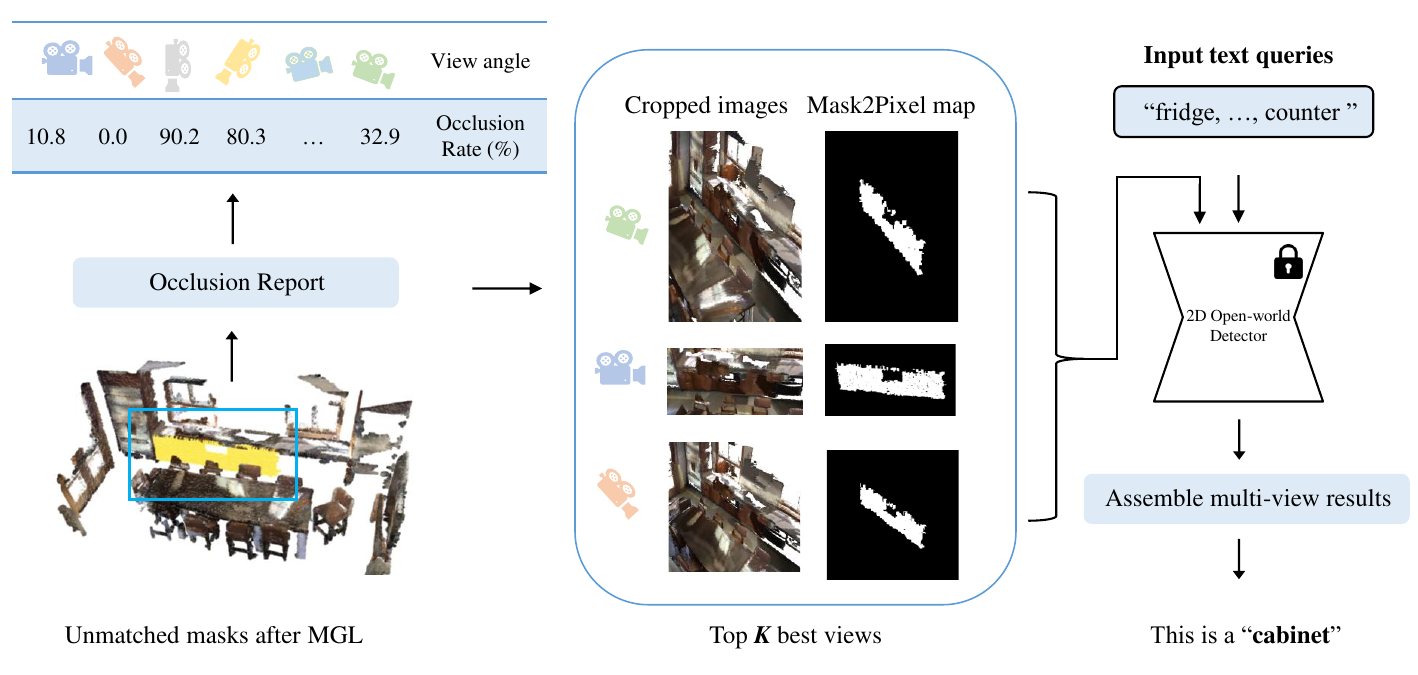}
 \centering
  \caption{\textbf{Illustration of Local Enforced Lookup.} The remaining masks from phase one first go through the \textit{Occlusion Report} module to select the best $K$ views. The selected images are cropped before being processed by the 2D detectors to encourage a classification result.}
 \label{fig: lel}
\end{figure*}

\noindent  After selecting the best view with \textit{Occlusion Report}, synthetic scene-level images are cropped to focus on a specific mask proposal and then reprocessed by 2D detectors. The results are also searched with the help of Mask2Pixel, in this case the binary mask, maps to form the final classification prediction for the mask, as shown in Figure \ref{fig: lel}.

\section{Implementation Details}
\label{section2}
\subsection*{Zero-shot Object Recognition}

Zero-shot object recognition results are obtained by employing the “Snap and Lookup'' modules to assign category names to the ground truth masks. The Snap module takes 24 images, and the Lookup module uses ODISE \cite{xu2023openvocabulary} to extract potentially interesting objects and assign labels to the masks. For masks that are not assigned a label or are assigned the wrong label, the top-1 classification is marked as a false negative.

\noindent Most other methods in this comparison use an object-centred rendering approach where a depth map or point cloud is projected into images for classification. Uni3D \cite{zhou2023uni3d} is pre-trained on a large amount of image and text pairs, as well as 3D shapes. OpenIns3D's scene-level image rendering, with enhanced results, proves to be much more effective in object recognition tasks.

\subsection*{Open-Vocabulary Instance Segmentation}

\noindent\textbf{Mask.} The Mask Proposal Module is built upon a lightweight version of Mask3D \cite{schult2023mask3d} with three decoder layers. For the mask quality scoring module, we set \(\lambda\) to 0.1 to down-weight zero IOU masks. The Mask Proposal Module is trained using the ADAM optimizer with a learning rate of 0.0003, and the one-cycle scheduler is applied. For the ScanNetv2, S3DIS, and STPLS3D experiments, the mask proposal module is trained on all-category class-agnostic masks to learn to propose masks. For Replica and ScanNet200, we followed OpenMask3D \cite{takmaz2023openmask3d} and used ScanNet200 pre-trained weights for mask proposals.

\noindent\textbf{Snap.} We captured 24 images of the scene, including 16 global, 4 corner, and 4 wide-angle images. We used the PyTorch3D Rasterization Renderer to render images. For all datasets, we captured images with dimensions of 1000 x 1000 for a great trade-off between speed and performance. Additionally, to avoid the occlusion effect caused by the ceiling, we discarded the top 0.3m points in the S3DIS and ScanNet datasets. As a result, the ceiling categories in the S3DIS dataset are completely discarded and assigned 0 in the AP results. For STPLS3D, the camera position is located 5m higher than the top of the scene to acquire a better view.

\noindent\textbf{Lookup.} During the \textit{Lookup} stage, we only assign a classification label to each mask if the results have been verified in at least two views. In the case of \textit{Local Enforced Lookup}, we crop the images using bounding boxes that are twice the size of the target masks. The cropped images are then fed into 2D detectors to refine the results. Mask2Pixel maps, in this case, binary maps, are used to accurately search for the detection results, as shown in Figure \ref{fig: lel}.

\subsection*{Open-vocabulary Object Detection}
Open-vocabulary object detection is carried out by converting the mask proposals into axis-aligned bounding boxes. We followed the same Snap and Lookup implementation details in Open-vocabulary Instance Segmentation Setting for bounding box understanding.

\section{Replacing Snap with RGB-D} 

While OpenIns3D primarily utilizes a 3D-input framework with the proposed "Mask-Snap-Lookup" pipeline, we also assess its performance in scenarios where 2D images are available, enabling comparison with other methods. The only modification made to the framework is the replacement of the Snap module with available RGB-D images. For evaluation, we selected two extensive datasets: ScanNet200\cite{rozenberszki2022language} and Replica\cite{straub2019replica}. The results are presented in Tables \ref{tab:scannet200_image} and \ref{tab:replica_image}. OpenIns3D with RGB-D excels on the Replica dataset while it performs competitively on the ScanNet200 dataset. Both methods use Yoloworld\cite{Cheng2024YOLOWorld} as the 2D detector. (\textit{updated on 1st Aug, 2024})

\begin{table}[!t]
\centering
\setlength{\tabcolsep}{5pt}
\caption{\textbf{3D Instance Segmentation Results on the ScanNet200 Validation Set.} OpenIns3D with RGBD images demonstrates competitive performance.}
\tablestyle{4.0pt}{1.4}
\scriptsize
\begin{tabular}{l|c|ccc|ccc}
\toprule
Model & use 2D & $\text{AP}_{\text{head}}$ & $\text{AP}_{\text{common}}$ & $\text{AP}_{\text{tail}}$ &  AP & AP$_{50}$  & AP$_{25}$ \\
\midrule
OpenScene (2D Fusion) \cite{peng2022openscene}   & \cmark & $13.4$ &$11.6$& $9.9$ &$11.7$ & $15.2$& $17.8$ \\ %
OpenScene (2D/3D Ens.) \cite{peng2022openscene}  & \cmark & $11.0$ &$3.2$& $1.1$  & $5.3$ & $6.7$& $8.1$ \\ %

OpenMask3D  & \cmark & ${17.1}$ &${14.1}$ & ${14.9}$   & ${15.4}$ & ${19.9}$& ${23.1}$  \\ 
    \rowcolor[gray]{.9}  OpenIns3D \tiny{with rgbd} &  \cmark & $\mathbf{19.2}$ &$\mathbf{14.2}$ & $\mathbf{14.2}$  & $\mathbf{15.9}$ & $\mathbf{20.6}$& $\mathbf{23.3}$   \\
    \hline
OpenScene (3D Distill)   & \xmark & $10.6$ & $2.6$ & $0.7$  & $4.8$ & $6.2$& $7.2$ \\ %
  \rowcolor[gray]{.9}  OpenIns3D &  \xmark & $\mathbf{16.0}$ &$\mathbf{6.5}$ & $\mathbf{4.2}$  & $\mathbf{8.8}$ & $\mathbf{10.3}$& $\mathbf{14.4}$   \\ %
      \bottomrule
\end{tabular}
\label{tab:scannet200_image}

\end{table}

\begin{table}[!t]
\centering
\setlength{\tabcolsep}{5pt}
\caption{\textbf{3D Instance Segmentation Results on the Replica.} OpenIns3D with RGBD images demonstrates competitive performance.} 
\tablestyle{4.0pt}{1.4}
\scriptsize

 \begin{tabular}{y{38mm}|x{6mm}x{8mm}x{8mm}x{8mm}}
    \toprule
    Model & 2D & AP & AP$_{50}$  & AP$_{25}$ \\
    \midrule
    OpenScene \cite{peng2022openscene} (2D Fusion)& \cmark &$10.9$ & $15.6$ & $17.3$ \\ %
    OpenScene \cite{peng2022openscene} (2D/3D Ens.)& \cmark & $8.2$ & $10.4$& $13.3$ \\ %
    OpenMask3D \cite{takmaz2023openmask3d} & \cmark  & ${13.1}$ & ${18.4}$& ${24.2}$  \\ %
    Open3DIS \cite{nguyen2023open3dis} & \cmark & ${18.5}$ & ${24.5}$& ${28.2}$  \\ %
 \rowcolor[gray]{.9}     OpenIns3D  \tiny{with rgbd}  & \cmark & \textbf{21.1} & \textbf{26.2} &\textbf{30.6}  \\

        \hline

    OpenScene \cite{peng2022openscene} (3D Distill) & \xmark  & $8.2$ & $10.5$& $12.6$ \\ %
  \rowcolor[gray]{.9}     OpenIns3D  & \xmark & \textbf{13.6} & \textbf{18.0} &\textbf{19.7}  \\
    \bottomrule
    \end{tabular}

\label{tab:replica_image}

\end{table}

\label{section7}

\section{Detailed Results}
\label{section3}

\subsection*{Open-vocabulary Instance Segmentation}
Tables \ref{tab:per_class_s3dis} and \ref{tab:per_class_scannet} provide the per-class results of OpenIns3D on the S3DIS and ScanNetv2 datasets. The novel (unseen) classes in PLA are highlighted in blue. Table \ref{tab:per_class_stpls3d} represents the per-category results for the STPLS3D dataset, compared with PointCLIP and PointCLIPV2.

In \textbf{S3DIS}, OpenIns3D consistently achieves high results in these novel classes. We attribute this to the high quality of 3D point data in S3DIS, which ensures favorable conditions for object detection in 2D Snap images. However, for classes like columns and beams, OpenIns3D struggles to produce desirable results. Ceiling results are marked as 0, as explained in the implementation detail section.

In \textbf{ScanNetv2}, the point cloud data quality is not very high. Consequently, the Snap output quality is limited, resulting in slightly lower performance.

In \textbf{STPLS3D}, OpenIns3D outperforms PointCLIP and PointCLIPV2 by a significant margin. This is as expected, as already demonstrated in the zero-shot classification tasks. However, the performance on very small objects, such as bikes, motorbikes, signs, and light poles, is not as strong. This is because the Snap module positions the camera at a high angle, resulting in a limited number of pixels available for these smaller objects.

\subsection*{Open-vocabulary Object Detection.}

We also present the object detection results at Table \ref{tab:ovod_3d} for all categories, followed by the OV-3DETIC categories. OpenIns3D shows strong performance across most categories.

\subsection*{Cross-domain analysis.} 

Table \ref{tab: per-class-cross-domain} presents the per-category results for the cross-domain OpenIns3D model, trained on S3DIS and tested on ScanNetv2. This table comparison is conducted to demonstrate the generalization capability of the mask module. 

\begin{table}[ht]
\begin{minipage}{\textwidth}
    \centering
    \caption{\textbf{Per-class Results of 3D Open-vocabulary Instance Segmentation on S3DIS AP50.} Performance on novel classes is marked in \colorbox{myblue2}{blue}.}
    \vspace{-3mm}
     \label{tab:per_class_s3dis}
       \tablestyle{0pt}{1.3}
       \scriptsize
        \begin{tabular}{x{17mm}|x{13mm}|x{7mm}x{7mm}x{7mm}x{7mm}x{7mm}x{7mm}x{7mm}x{7mm}x{7mm}x{7mm}x{7mm}x{7mm}}
            \bottomrule[1pt]
            Methods & Partition & \rotatebox[origin=c]{90}{ ceiling } & \rotatebox[origin=c]{90}{floor} & \rotatebox[origin=c]{90}{wall} & \rotatebox[origin=c]{90}{beam} & \rotatebox[origin=c]{90}{column} & \rotatebox[origin=c]{90}{window} & \rotatebox[origin=c]{90}{door} & \rotatebox[origin=c]{90}{table} & \rotatebox[origin=c]{90}{chair} & \rotatebox[origin=c]{90}{sofa} & \rotatebox[origin=c]{90}{ bookcase } & \rotatebox[origin=c]{90}{board} \\
            \hline
            \multirow{2}{*}{PLA \cite{ding2022language}} & B8/N4 & 89.5 & 100.0 & 50.8 & 00.0 & 35.3 & {\cellcolor{myblue2}36.2} & 60.5 & {\cellcolor{myblue2}00.1} & 84.6 & {\cellcolor{myblue2}01.9} & {\cellcolor{myblue2}00.8} & 59.4 \\
            & B6/N6 & 89.5 & {\cellcolor{myblue2}60.2} & 17.9 & 00.0 & 41.5 & {\cellcolor{myblue2}10.2} & {\cellcolor{myblue2}02.1} & {\cellcolor{myblue2}00.6} & 86.2 & 45.1 & {\cellcolor{myblue2}00.1} & {\cellcolor{myblue2}02.2} \\ \hline
            OpenIns3D & --/N12 & \cellcolor{myblue2}  00.0 & \cellcolor{myblue2} 84.4 &\cellcolor{myblue2} 29.0 &\cellcolor{myblue2} 00.0 &\cellcolor{myblue2} 00.0 &\cellcolor{myblue2} 62.6 &\cellcolor{myblue2} 25.2 &\cellcolor{myblue2} 25.5 &\cellcolor{myblue2} 52.0 &\cellcolor{myblue2} 60.0 &\cellcolor{myblue2} 00.0 &\cellcolor{myblue2} 00.0\\
            \toprule[1pt]
        \end{tabular}

     \caption{\textbf{Per-class Results of 3D Open-vocabulary Instance Segmentation on ScanNet AP50.} Performance on novel classes is marked in \colorbox{myblue2}{blue}.}
    \begin{small}
      \resizebox{\textwidth}{!}{
        \setlength{\tabcolsep}{1.3mm}{
        \begin{tabular}{c|c|ccccccccccccccccccc}
            \bottomrule[1pt]
            Methods & Partition &  \rotatebox[origin=c]{90}{cabinet} & \rotatebox[origin=c]{90}{bed} & \rotatebox[origin=c]{90}{chair} & \rotatebox[origin=c]{90}{sofa} & \rotatebox[origin=c]{90}{table} & \rotatebox[origin=c]{90}{door} & \rotatebox[origin=c]{90}{window} & \rotatebox[origin=c]{90}{ bookshelf } & \rotatebox[origin=c]{90}{picture} & \rotatebox[origin=c]{90}{counter} & \rotatebox[origin=c]{90}{desk} & \rotatebox[origin=c]{90}{curtain} & \rotatebox[origin=c]{90}{fridge} & \rotatebox[origin=c]{90}{shower c.} & \rotatebox[origin=c]{90}{toilet} & \rotatebox[origin=c]{90}{sink} & \rotatebox[origin=c]{90}{bathtub} \\
            \hline
            
            \multirow{3}{*}{PLA \cite{ding2022language}} & B13/N4  & 50.5 & 77.0 & 82.9 & {\cellcolor{myblue2}43.4} & 75.4 & 49.0 & 46.0 & {\cellcolor{myblue2}43.7} & 46.5 & 33.7 & {\cellcolor{myblue2}23.2} & 54.1 & 49.6 & 56.0 & {\cellcolor{myblue2}97.8} & 47.5 & 85.8 \\
            & B10/N7   & 53.7 & {\cellcolor{myblue2}62.7} & {\cellcolor{myblue2}11.2} & 70.5 & {\cellcolor{myblue2}27.2} & 47.7 & 45.7 & {\cellcolor{myblue2}30.0} & {\cellcolor{myblue2}01.5} & 39.9 & 40.8 & 50.6 & 68.6 & 84.6 & 92.9 & {\cellcolor{myblue2}24.6} & {\cellcolor{myblue2}00.0} \\
            & B8/N9   & 45.1 & 77.4 & 82.2 & 84.2 & 74.2 & 48.9 & 51.0 & {\cellcolor{myblue2}30.0} & {\cellcolor{myblue2}00.5} & {\cellcolor{myblue2}02.1} & {\cellcolor{myblue2}16.8} & 44.9 & {\cellcolor{myblue2}28.3} & {\cellcolor{myblue2}35.1} & {\cellcolor{myblue2}94.3} & {\cellcolor{myblue2}16.6} & {\cellcolor{myblue2}00.0} \\ \hline
            OpenIns3D & --/N17   & \cellcolor{myblue2} 24.3 & \cellcolor{myblue2} 52.5 &\cellcolor{myblue2} 75.7 & \cellcolor{myblue2} 61.6 & \cellcolor{myblue2} 40.6 & \cellcolor{myblue2} 39.7 & \cellcolor{myblue2}45.5 & \cellcolor{myblue2} 54.8 & \cellcolor{myblue2} 0.5 &\cellcolor{myblue2} 33.5 &\cellcolor{myblue2} 16.7 & \cellcolor{myblue2} 48.1 & \cellcolor{myblue2} 18.5 & \cellcolor{myblue2} 4.3 &\cellcolor{myblue2} 50.1 &\cellcolor{myblue2} 16.8 &\cellcolor{myblue2} 7.6\\
            \toprule[1pt]
        \end{tabular}}}
            \label{tab:per_class_scannet}
    \end{small}
\end{minipage}
\end{table}

\begin{table}[ht]
\centering
     \caption{\textbf{Per-class Results of 3D Open-vocabulary Instance Segmentation on STPLS3D AP50.}  All models are tested in a zero-shot manner.}
     \label{tab:per_class_stpls3d}
\tablestyle{0.7pt}{2.0}
    \tiny
    \begin{tabular}{x{20mm}|x{7mm}|x{7mm}x{7mm}x{7mm}x{7mm}x{7mm}x{7mm}x{7mm}x{7mm}x{7mm}x{7mm}x{7mm}x{7mm}}\toprule
Methods & mean &
\rotatebox[origin=c]{90}{building}  &\rotatebox[origin=c]{90}{veg}  &\rotatebox[origin=c]{90}{vehicle}  &\rotatebox[origin=c]{90}{truck}  &\rotatebox[origin=c]{90}{aircraft}  &\rotatebox[origin=c]{90}{mil-veh} &\rotatebox[origin=c]{90}{bike}  &\rotatebox[origin=c]{90}{motorbike}  &\rotatebox[origin=c]{90}{lightpole}  &\rotatebox[origin=c]{90}{signs}  &\rotatebox[origin=c]{90}{clutter}  &\rotatebox[origin=c]{90}{fence}   \\\midrule
PointCLIP \cite{zhu2022pointclip} & 2.7 &15.3 &0.4 &10.2 &06.6 &00.0 &00.0 &00.0 &00.0 &00.0 &00.0 &00.0 &00.0  \\
PointCLIPV2 \cite{zhang2022pointclip}& 3.2 &20.3 &0.2 &12.3 &5.8 &00.0 &00.0 &00.0 &00.0 &00.0 &00.0 &00.0 &00.0  \\ \hline
 \rowcolor[gray]{.9}  OpenIns3D & \textbf{14.1 }&\textbf{40.4} &\textbf{01.2} &\textbf{54.2} &\textbf{24.2} &\textbf{30.0} &\textbf{05.5} &\textbf{02.1} &\textbf{03.0} &00.0 &00.0 &00.0 &\textbf{08.3}  \\
\bottomrule
\end{tabular}

\begin{small}
\caption{Detailed results on 3D open vocabulary object detection.  We present all results on ScanNet20, followed by OV-3DETIC \cite{lu2023open}.}
\tablestyle{0.7pt}{2.0}
\tiny
\begin{tabular}{l|c|x{4.6mm}x{4.6mm}x{4.6mm}x{4.6mm}x{4.6mm}x{4.6mm}x{4.6mm}x{4.6mm}x{4.6mm}x{4.6mm}x{4.6mm}x{4.6mm}x{4.6mm}x{4.6mm}x{4.6mm}x{4.6mm}x{4.6mm}x{4.6mm}x{4.6mm}x{4.6mm}}\toprule
\textbf{Methods} & \textbf{mean} 
& \rotatebox[origin=c]{90}{toilet} 
& \rotatebox[origin=c]{90}{bed} 
& \rotatebox[origin=c]{90}{chair} 
& \rotatebox[origin=c]{90}{sofa} 
& \rotatebox[origin=c]{90}{dresser} 
& \rotatebox[origin=c]{90}{table} 
& \rotatebox[origin=c]{90}{cabinet} 
& \rotatebox[origin=c]{90}{bookshelf} 
& \rotatebox[origin=c]{90}{pillow} 
& \rotatebox[origin=c]{90}{sink} 
& \rotatebox[origin=c]{90}{bathtub}
& \rotatebox[origin=c]{90}{fridge}
& \rotatebox[origin=c]{90}{desk}
& \rotatebox[origin=c]{90}{nightstand}
& \rotatebox[origin=c]{90}{counter} 
& \rotatebox[origin=c]{90}{door} 
& \rotatebox[origin=c]{90}{curtain} 
& \rotatebox[origin=c]{90}{box} 
& \rotatebox[origin=c]{90}{lamp} 
& \rotatebox[origin=c]{90}{bag}  \\ \midrule
3DETIC \cite{zhou2022detecting} & 15.0 & 53.3 & 24.9 & 15.8 & 31.4 & \textbf{11.5} & 09.1 & 02.1 & 09.4 & 17.0 & 29.2 & 27.5 & 20.0 & 13.7 & 00.0 & 00.0 & 00.0 & 17.7 & 04.8 & 03.0 & \textbf{09.5} \\
CLIP-3D  \cite{clip}& 12.7 & 44.8 & 23.8 & 17.5 & 12.6 & 04.9 & 13.2 & 01.9 & 04.0 & 11.4 & 17.6 & 32.2 & 14.9 & 11.4 & 02.4 & 00.5 & 14.5 & 08.6 & \textbf{07.5} & 05.1 & 04.7 \\
OV-3DETIC \cite{lu2023open} & 18.8 & 57.3 & 42.3 & 27.1 & \textbf{31.5} & 08.2 & 14.2 & 03.0 & 05.6 & \textbf{23.0} & 31.6 & \textbf{56.3} & 11.0 & 19.7 & 00.8 & 00.3 & 09.6 & 10.5 & 03.8 & 02.1 & 02.7 \\ \hline
\rowcolor[gray]{.9}  OpenIns3D & \textbf{37.1} & \textbf{79.5} & \textbf{70.5} & \textbf{76.9} & 15.8 & 00.0 & \textbf{53.1} & \textbf{40.1} & \textbf{41.2} & 07.1 & \textbf{53.1} & 14.3 & \textbf{32.1} & \textbf{29.1} & \textbf{04.8 }& \textbf{55.6} & \textbf{40.4} & \textbf{41.1} & 02.6 & \textbf{48.0} & 06.2 \\

\bottomrule
\end{tabular}
\label{tab:ovod_3d}
\end{small}

\end{table}

\begin{table}[!htp]\centering
\caption{\textbf{Cross-domain Analysis of OpenIns3D on OVOD on ScanNetv2 AP25.} OpenIns3D achieves competitive results on the cross-domain dataset, even on categories at are not available on the training dataset, highlighted in \colorbox{myblue2}{blue}. Compared with other SOTA models on OVOD, cross-domain OpenIns3D still has competitive performance. 
MPM-SC: MPM trained on ScanNetv2; MPM-S3: MPM trained on S3DIS. 
}\label{tab: per-class-cross-domain}
\small \setlength{\tabcolsep}{2mm}{
 \resizebox{\textwidth}{!}{
\begin{tabular}{l|c|cccccccccccccccccc}\toprule
Methods & \rotatebox[origin=c]{90}{BBox Prop} & \rotatebox[origin=c]{90}{cabinet} & \rotatebox[origin=c]{90}{bed} & \rotatebox[origin=c]{90}{chair} & \rotatebox[origin=c]{90}{sofa} & \rotatebox[origin=c]{90}{table} & \rotatebox[origin=c]{90}{door} & \rotatebox[origin=c]{90}{window} & \rotatebox[origin=c]{90}{bookshelf } & \rotatebox[origin=c]{90}{picture} & \rotatebox[origin=c]{90}{counter} & \rotatebox[origin=c]{90}{desk} & \rotatebox[origin=c]{90}{curtain} & \rotatebox[origin=c]{90}{fridge} & \rotatebox[origin=c]{90}{shower c.} & \rotatebox[origin=c]{90}{toilet} & \rotatebox[origin=c]{90}{sink} & \rotatebox[origin=c]{90}{bathtub}
\\\midrule
OpenIns3D &MPM-SC &17.1 &57.5 &74.5 &59.2 &36.9 &29.3 &47.5 &26.4 &0.0 &31.1 &32.2 &55.4 &39.1 &0.0 &57.4 &42.1 &6.6 \\
OpenIns3D & MPM-S3&\cellcolor{myblue2}16.1 &\cellcolor{myblue2}43.5 &45.7 &41.8 &28.6 &17.7 &18.3 &31.9 &\cellcolor{myblue2}1.2 &\cellcolor{myblue2}1.0 &\cellcolor{myblue2}29.3 &\cellcolor{myblue2}23.1 &\cellcolor{myblue2}20.1 &\cellcolor{myblue2}8.0 &\cellcolor{myblue2}63.6 &\cellcolor{myblue2}{16.4} &\cellcolor{myblue2}1.7 \\ \hline
\textit{SOTA models} \\

PointCLIP &P-3DE &6.0 &4.8 &45.2 &4.8 &7.4 &4.6 &2.2 &- &- &1.0 &4.0 &- &- &- &- &13.4 &6.5 \\
PointCLIPV2  &P-3DE &19.3 &21.0 &61.9 &15.6 &23.8 &13.2 &17.4 &- &- &12.4 &21.4 &- &- &- &- &14.5 &16.8 \\
OV-3DET &P-2DE &3.0 &42.3 &27.1 &31.5 &14.2 &9.6 &- &5.6 &- &0.3 &19.7 &10.5 &11.0 &- &57.3 &31.6 &56.3 \\
\bottomrule
\end{tabular}}}
\end{table}

 \begin{figure*}[tp]
 \centering
 \includegraphics[width=1.0\textwidth]{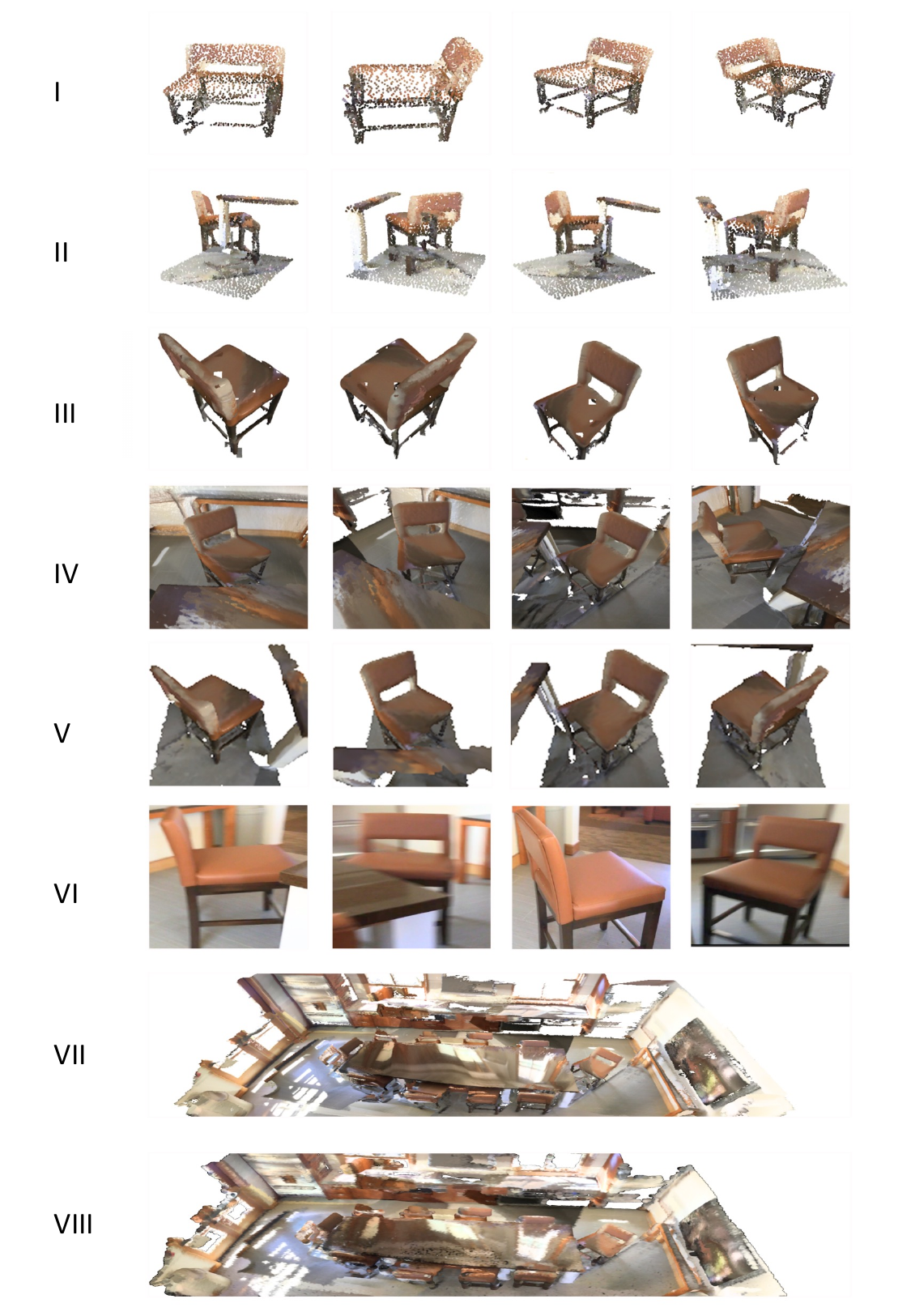}
 \centering
  \caption{\textbf{Visualization of Attempts Made to Generate 2D Images from 3D.} I: LAR-point projection; II: LAR-point-bg-project; III: Mesh rendering; IV Mesh-in-scene Rendering; V: Mesh-bg-Rendering; VI: Cropped from Original 2D images; VII: Scene Level Rendering from Mesh; VIII: Scene Level Rendering from Point. Performance can be found in Table \ref{tab：render attempts}.}
\label{fig: render_evo}
\end{figure*}

\section{Other Attempts for Image Generation}
\label{section4}

\begin{table}[!htp]\centering
\caption{\textbf{Evolution of \textit{Snap} and \textit{Lookup} Module.} The corresponding image visualization is shown in Figure \ref{fig: render_evo}. Scene-level rendering not only requires fewer images but also achieves superb results when compared to other pre-mask levels of rendering. \scriptsize{*}: The image sizes of VI are not fixed as it depends on the size of the mask area on the original images.}
\scriptsize

\begin{tabular}{c|c|ccccc|cc}\toprule

Idx &Methods & Job intensity &Imgs needed &use 2D & Img size&2D backbone &AP50 &AP25 \\\midrule
I &LAR-point projection  &per mask &250 &\xmark &$128^2$ &CLIP &5.3 &8.6 \\
II &LAR-point-bg-projection  &per mask &250 &\xmark &$128^2$ &CLIP  &6.3 &10.5 \\
III &mesh-rendering &per mask &250 &\xmark &$128^2$ &CLIP  &6.8 &7.2 \\
IV &mesh-scene-rendering&per mask &250 &\xmark &$128^2$ &CLIP &6.7 &7.3 \\
V &mesh-bg-rendering &per mask &250 &\xmark &$128^2$ &CLIP  &4.3 &5.3 \\
VI &crop-original2d &per mask &250 &\cmark & $-^*$ &CLIP  &24.3 &29.6 \\ \hline
  \rowcolor[gray]{.9} VII &scene-mesh-rendering & per scene &8 &\xmark &$1000^2$ &ODISE  &28.7 &38.9 \\
   \rowcolor[gray]{.9} VIII &scene-point-rendering & per scene &8 &\xmark &$1000^2$ &ODISE &21.5 &33.6 \\
\bottomrule
\end{tabular}

\label{tab：render attempts}
\end{table}

\label{appendx:rendering_ablation}
Figures \ref{fig: render_evo} and Table \ref{tab：render attempts} illustrate the alternative 2D image rendering approaches we explored before concluding that synthetic scene-level images offer the optimal solution. We document the process here for future reference.

\noindent \textbf{Attempts I, II:} Inspired by the success of LAR \cite{bakr2022look}, we positioned the camera around the object and projected point clouds to generate multi-view images for each mask. However, these approaches produced images beyond the recognition capability of the CLIP model, especially for masks that were not well-segmented or reconstructed, leading to unsatisfactory results.

\noindent \textbf{Attempts III, IV, V:} We redirected our attention to the mesh of the scene, using rasterization rendering methods rather than simple point projecting. Although these methods brought some improvement, they still proved to have strong limitations when the mask proposed was not perfect. Masks of undesired quality made up a large portion of the masks, making this an inadequate solution. Moreover, per-mask rendering required a significant amount of time, making it impractical for deployment.

\noindent \textbf{Attempt VI:} We then experimented with using original images and cropping out masks in the images for evaluation (VI). We believed this would offer the best quality of images, making them most likely to be recognizable with 2D models. We used \textit{Occlusion Reports} methods to select the top $K$ views from all frames and crop out mask pixels with an enlarged bounding box. This approach achieved notable performance, primarily due to the high quality of 2D images. However, we ultimately abandoned this approach due to concerns about its applicability in general scenarios.

\noindent \textbf{Attempts VII, VIII:} Shifting our focus to scene-level rendering, our model began to produce high-quality results. By observing all broken instances from a distance and incorporating a large amount of contextual information, objects became clear and recognizable. As a result, the scene-level images had a small domain gap with the images used to train 2D Vision-Language models.

\section{Limitations and Future Work}

\label{section5}

There are some limitations of OpenIns3D that need further investigation in future studies.

\begin{itemize}

\item Reliance on ground truth instance masks: Similar to SAM \cite{kirillov2023segment}, OpenIns3D still relies on ground truth mask supervision. While it does prove to have the capability to generalize masks that have never been seen before, providing a vast amount of class-agnostic masks can be helpful. Approaches like UnScene3D \cite{rozenberszki2023unscene3d}, Segment3D \cite{Huang2023Segment3D} might serve as alternative methods for mask proposal, linking it with \textit{Snap} and the \textit{Lookup} module for open-vocabulary understanding. This requires further investigation.

\item Limited performance in semantic segmentation: OpenIns3D heavily relies on filtering to refine the mask proposals, discarding masks with low quality directly. While this approach benefits instance segmentation by reducing false positive instances, it may limit its performance in semantic segmentation. We have also calculated the semantic segmentation results of OpenIns3D on four categories, as reported by OpenScene \cite{peng2022openscene}, as shown in Table \ref{tab: semantic-seg}. Our method still exhibits a gap compared to OpenScene in terms of semantic segmentation.

\item Small object performance: The performance of OpenIns3D is ultimately closely linked to the quality of the point cloud itself. Masks that are very small or made of sparse point clouds would be difficult to recognize in the rendered images, as they either occupy a small portion of the image pixels or are too fragmented to be detected by the 2D models. 
\end{itemize}

\vspace{-3mm}

\begin{table}[t]\centering
\caption{\textbf{Comparison with OpenScene and other Frameworks on Semantic Segmentation.} Our framework prioritises mask quality and sacrifice overall semantic segmentation results. }
\small
\resizebox{\textwidth}{!}{
\begin{tabular}{lrrrrrrrrrrr}\toprule
Semantic Seg. &\multicolumn{5}{c}{mIoU} &\multicolumn{5}{c}{mAcc} \\\midrule
Methods &Bookshelf &Desk &Sofa &Toilet &Mean &Bookshelf &Desk &Sofa &Toilet &Mean \\
3DGenZ \cite{michele2021generative} &6.3 &3.3 &13.1 &8.1 &7.7 &13.4 &5.9 &5.9 &26.3 &12.9 \\
MSeg Voting \cite{mseg}&47.8 &40.3 &56.5 &68.8 &53.3 &50.1 &67.7 &67.7 &81.0 &66.6 \\
OpenScene-LSeg \cite{peng2022openscene} &\textbf{67.1} &\textbf{46.4} &60.2 &\textbf{77.5} &\textbf{62.8} &\textbf{85.5} &\textbf{69.5} &69.5 &90.0 &78.6 \\
OpenScene-OpenSeg \cite{peng2022openscene} &64.1 &27.4 &49.6 &63.7 &51.2 &73.7 &73.4 &73.4 &95.3 &79.0 \\ \hline
OpenIns3D &54.8 &16.7 &\textbf{61.6} &50.6 &45.9 &59.0 &32.3 &\textbf{76.7} &79.8 &61.9 \\
\bottomrule
\end{tabular}}
\label{tab: semantic-seg}
\end{table}

\section{Visualization}

\label{section6}
\begin{figure*}[t]
 \centering
 \includegraphics[width=\textwidth]{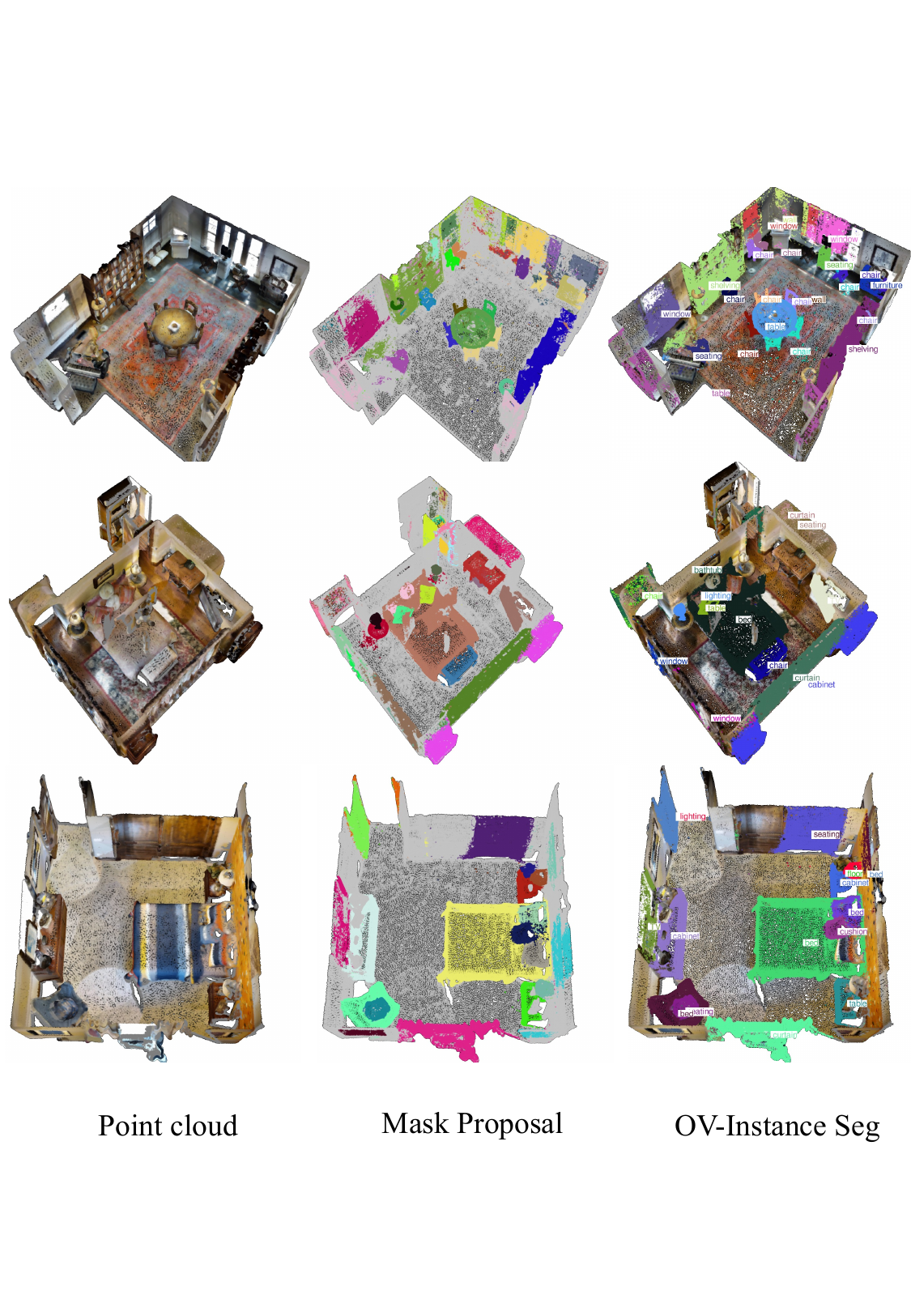}
 \centering
  \caption{\textbf{Zero-shot Open world Instance Segmentation on Matterport3D.}}
 \label{fig: mattport3d_zero}
 \vspace{-4mm}
\end{figure*}

\begin{figure*}[t]
 \centering
 \includegraphics[width=\textwidth]{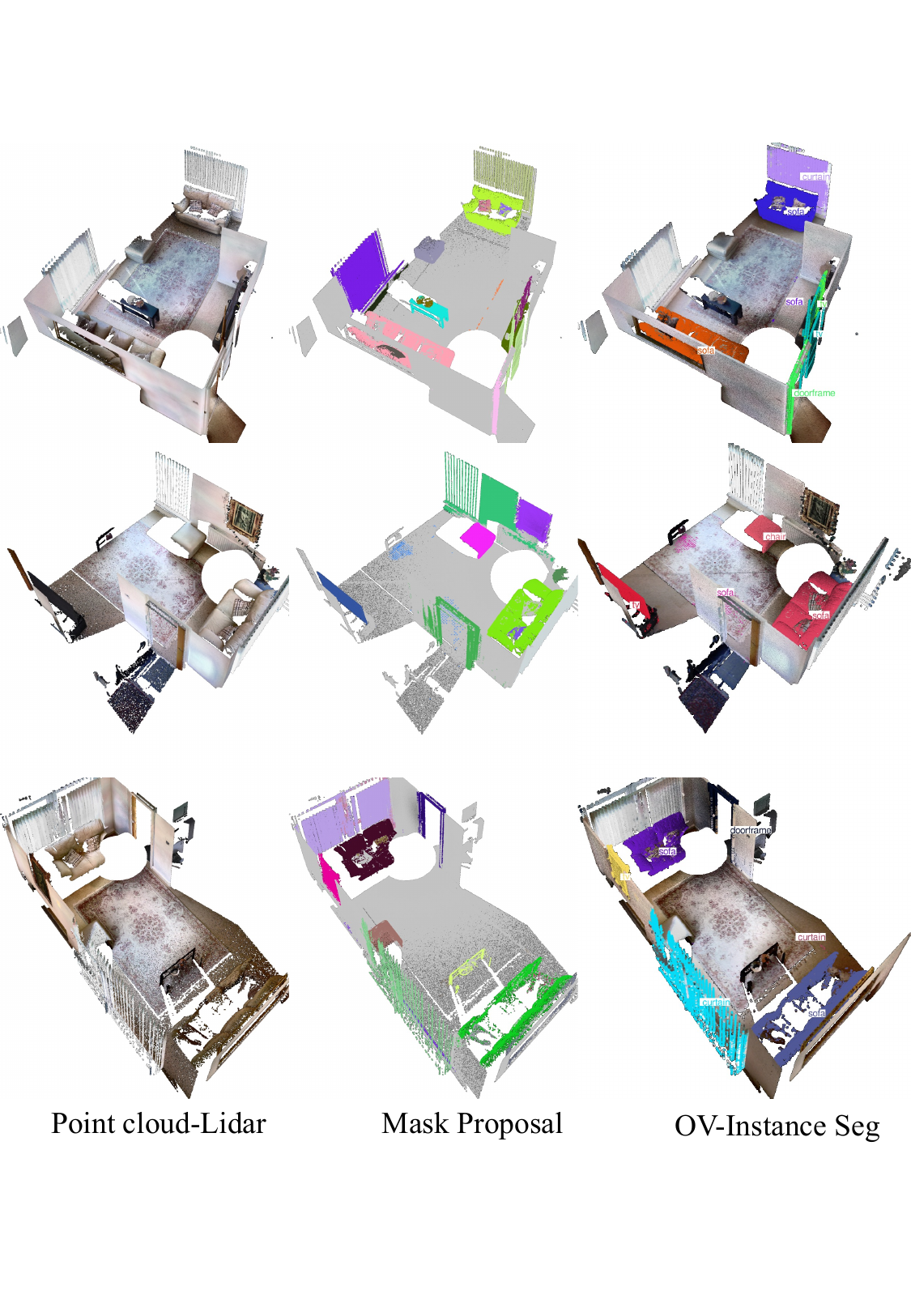}
 \centering
  \caption{\textbf{Zero-shot Open world Instance Segmentation on ArkitScene-Lidar.}}
 \label{fig: lidar_zero}
 \vspace{-4mm}
\end{figure*}

\noindent\textbf{Zero-shot performance on other dataset.} OpenIns3D is able of deploying on any colored 3D scans, regardless of the availability of 2D image counterparts. To demonstrate this, we provide some demos deploying OpenIns3D on Lidar-based datasets like ArkitScene Lidar, whose 2D images are not available, and Mattport3D, which has a different style of indoor rooms. The results are shown in Figure \ref{fig: lidar_zero} and Figure \ref{fig: mattport3d_zero}. We present both mask proposals (pre-trained on ScanNet 200) as well as the final detection results.

\newpage

\begin{figure*}[!ht]
 \vspace{-8mm}
 \centering
 \includegraphics[width=\textwidth]{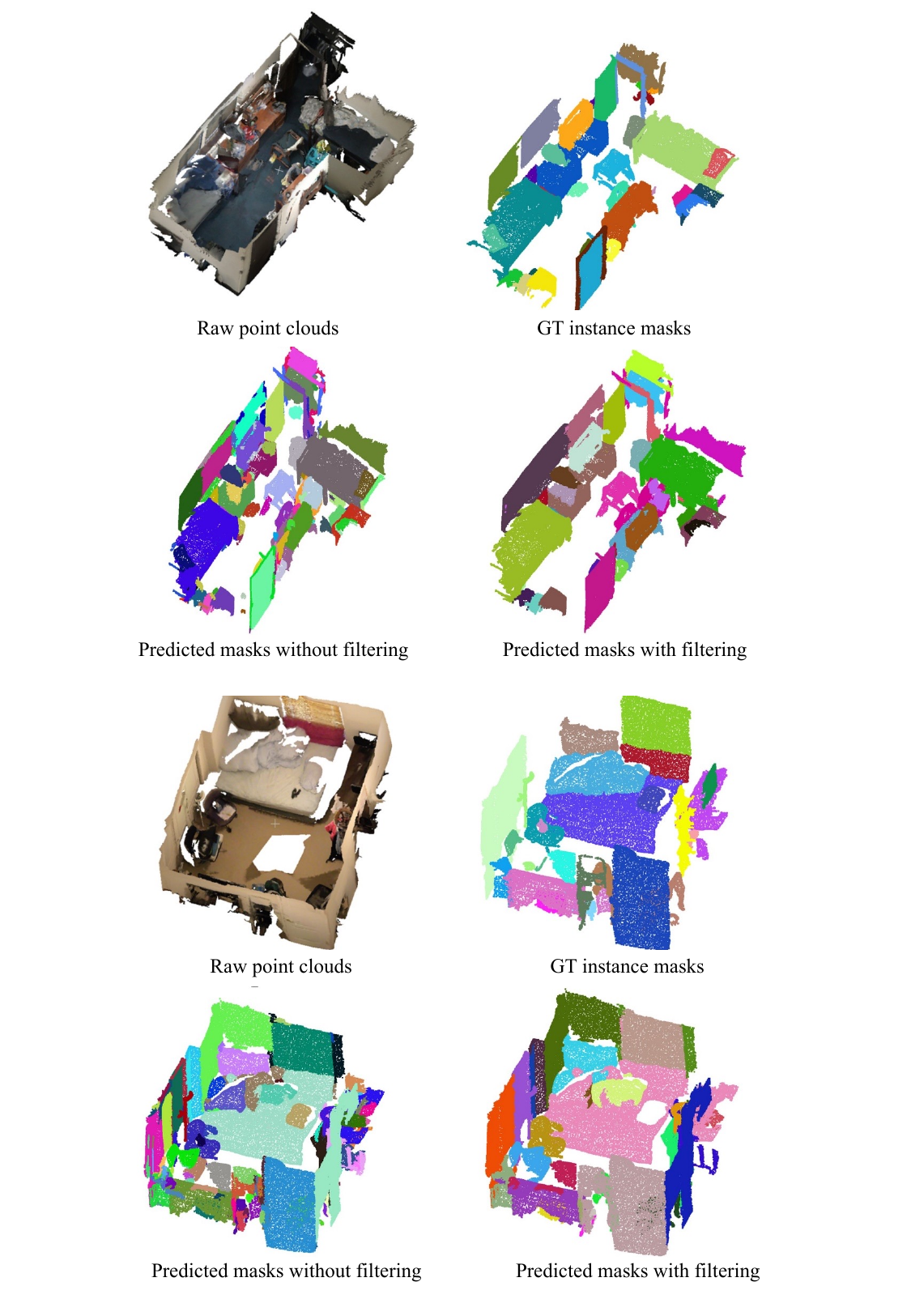}
 \centering
  \caption{\textbf{Qualitative Evaluation of the Mask Proposals.} Our class-label-free approach produces high-quality masks that closely resemble the ground truth. Additionally, the incorporation of \textit{Mask Scoring} and \textit{Mask Filtering} further enhances the overall quality of the masks.}
 \label{fig: mask_1}
 \vspace{-4mm}
\end{figure*}

\begin{figure*}[!ht]
 \vspace{-8mm}
 \centering
 \includegraphics[width=0.9\textwidth]{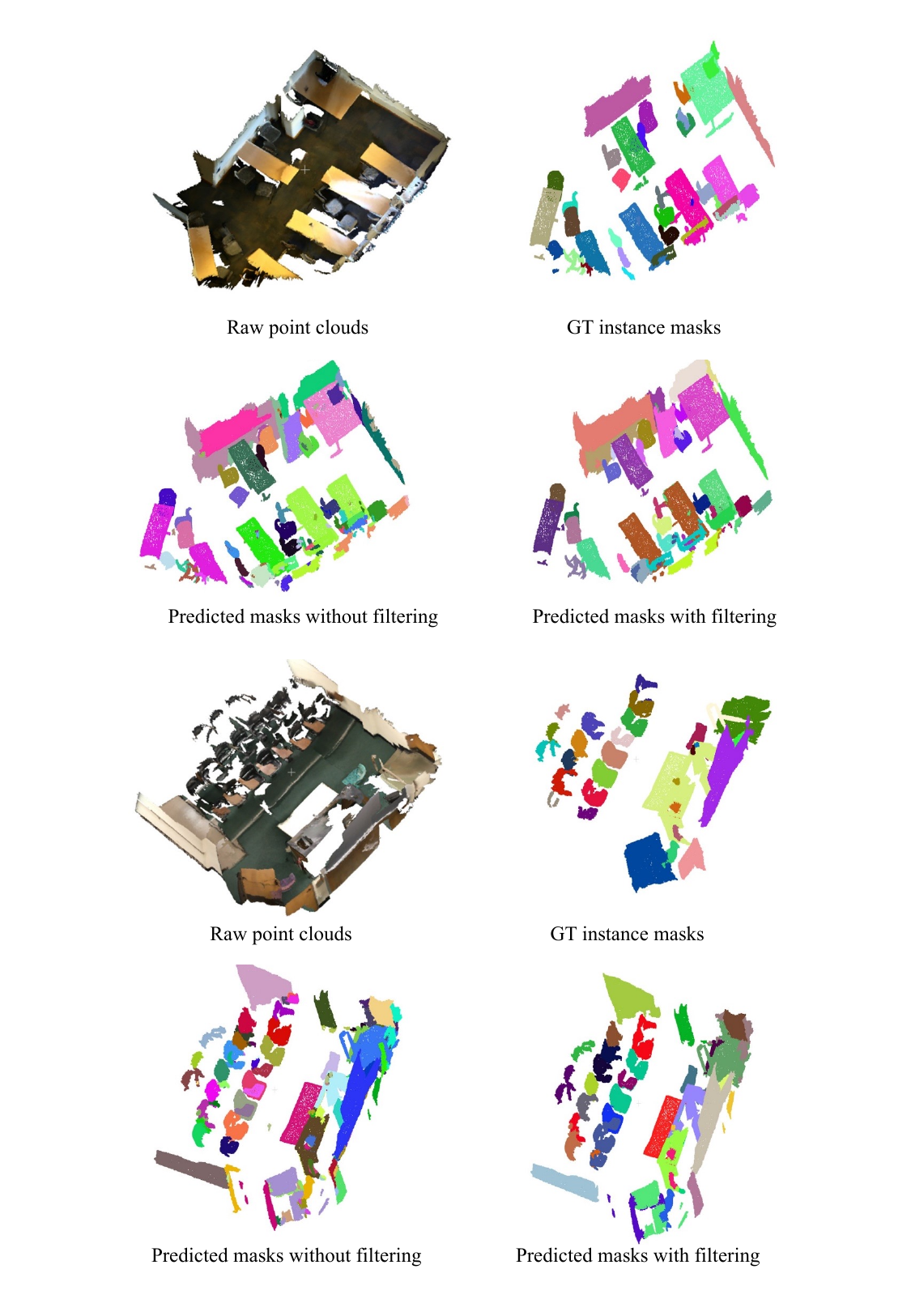}
 \centering
  \caption{\textbf{Qualitative Evaluation of the Mask Proposals.} Our class-label-free approach produces high-quality masks that closely resemble the ground truth. Additionally, the incorporation of \textit{Mask Scoring} and \textit{Mask Filtering} further enhances the overall quality of the masks.}
 \label{fig: mask_2}
 \vspace{-4mm}
\end{figure*}

\noindent\textbf{Mask proposal.}
\label{Section_3}
Figure \ref{fig: mask_1} and \ref{fig: mask_2} present a qualitative evaluation of the mask proposal module. The learned mask proposals exhibit great similarity to the ground truth masks, often capturing additional unlabeled masks. This demonstrates the effectiveness of our class-label-free learning scheme in producing high-quality class-agnostic mask proposals. Moreover, through the application of \textit{Mask Scoring} and \textit{Mask Filtering} techniques, we are able to connect fragmented or fragile masks, resulting in a substantial improvement in mask quality. These advancements provide a strong foundation for the Snap and Lookup understanding scheme.

\begin{figure*}[t!]
 \centering
 \includegraphics[width=1.0\textwidth]{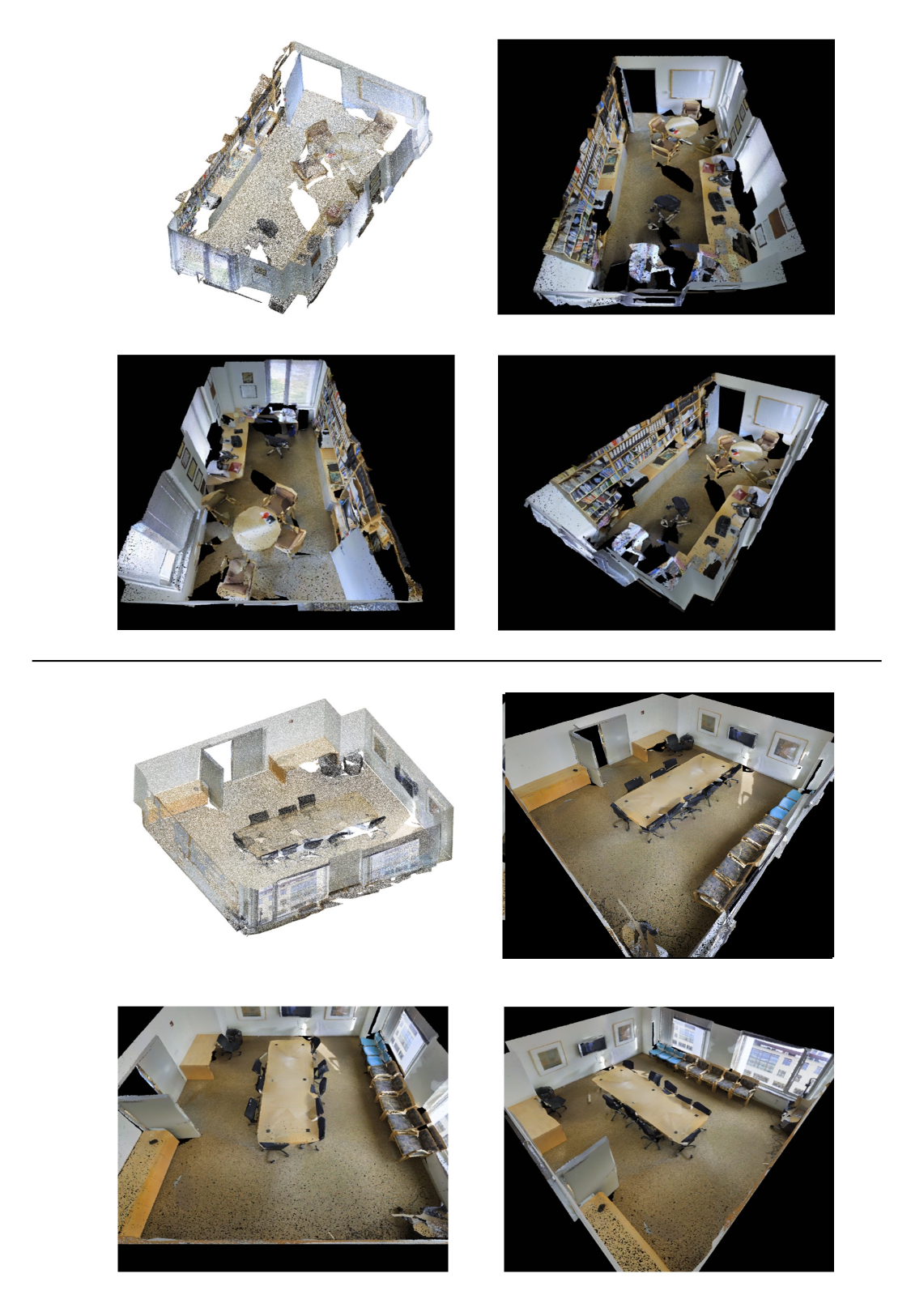}
 \centering
  \caption{\textbf{Synthetic Scene-level Images of S3DIS Generated by \textit{Snap}.} The first image is the original spare point cloud, and the following three images are outcomes of the \textit{Snap} module.}
 \label{fig: snap_s3dis}
\end{figure*}

\begin{figure*}[t!]
 \centering
 \includegraphics[width=1.0\textwidth]{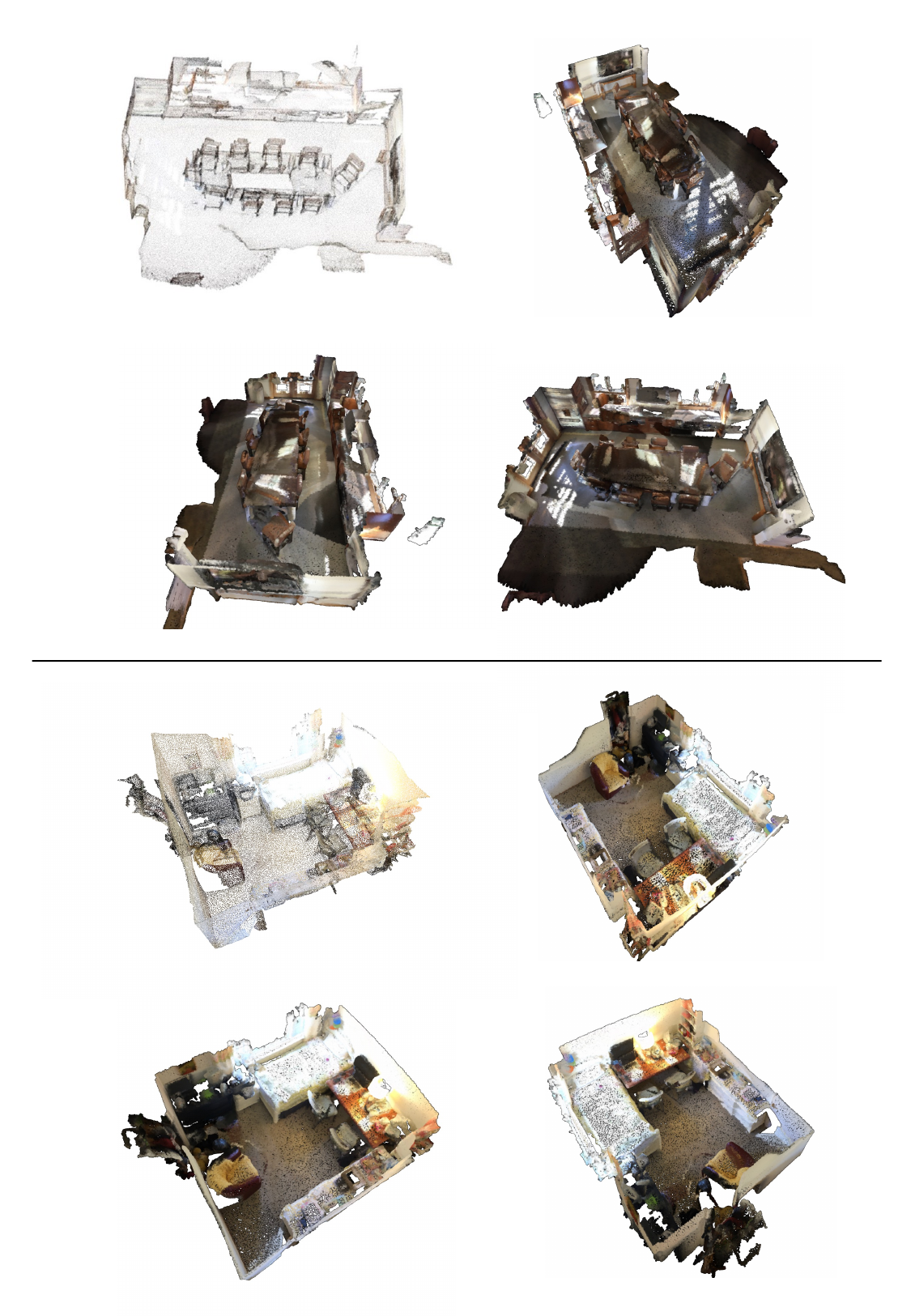}
 \centering
  \caption{\textbf{Synthetic Scene-level Images of ScanNetv2 Generated by \textit{Snap}.} The first image is the original spare point cloud, and the following three images are outcomes of the \textit{Snap} module.}
 \label{fig: snap_scannet}
\end{figure*}

\begin{figure*}[t!]
 \centering
 \includegraphics[width=1.0\textwidth]{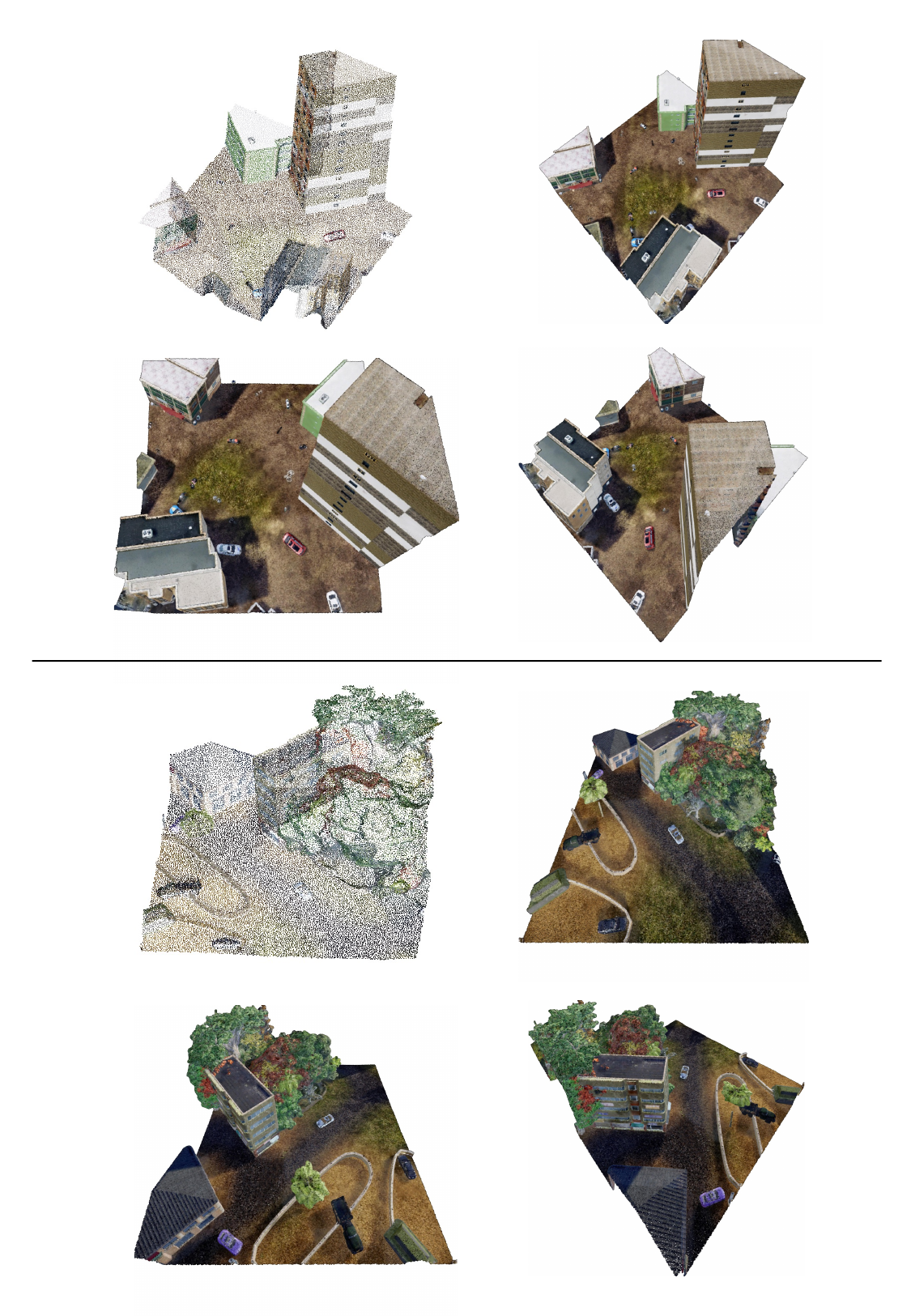}
 \centering
  \caption{\textbf{Synthetic Scene-level Images of STPLS3D Generated by \textit{Snap}.} The first image is the original spare point cloud, and the following three images are outcomes of the \textit{Snap} module.}
 \label{fig: snap_stpls3d}
\end{figure*}

\noindent\textbf{Snap visualization.}
Figure \ref{fig: snap_s3dis}, \ref{fig: snap_scannet} and \ref{fig: snap_stpls3d} demonstrate the capability of the Snap module, we present mostly the global snap. With the proposed pose and intrinsic optimization scheme, the Snap module is capable of generating decent-quality images from point clouds, regardless of whether the dataset is indoor or outdoor.

\begin{figure*}[t!]
 \centering
 \includegraphics[width=1.0\textwidth]{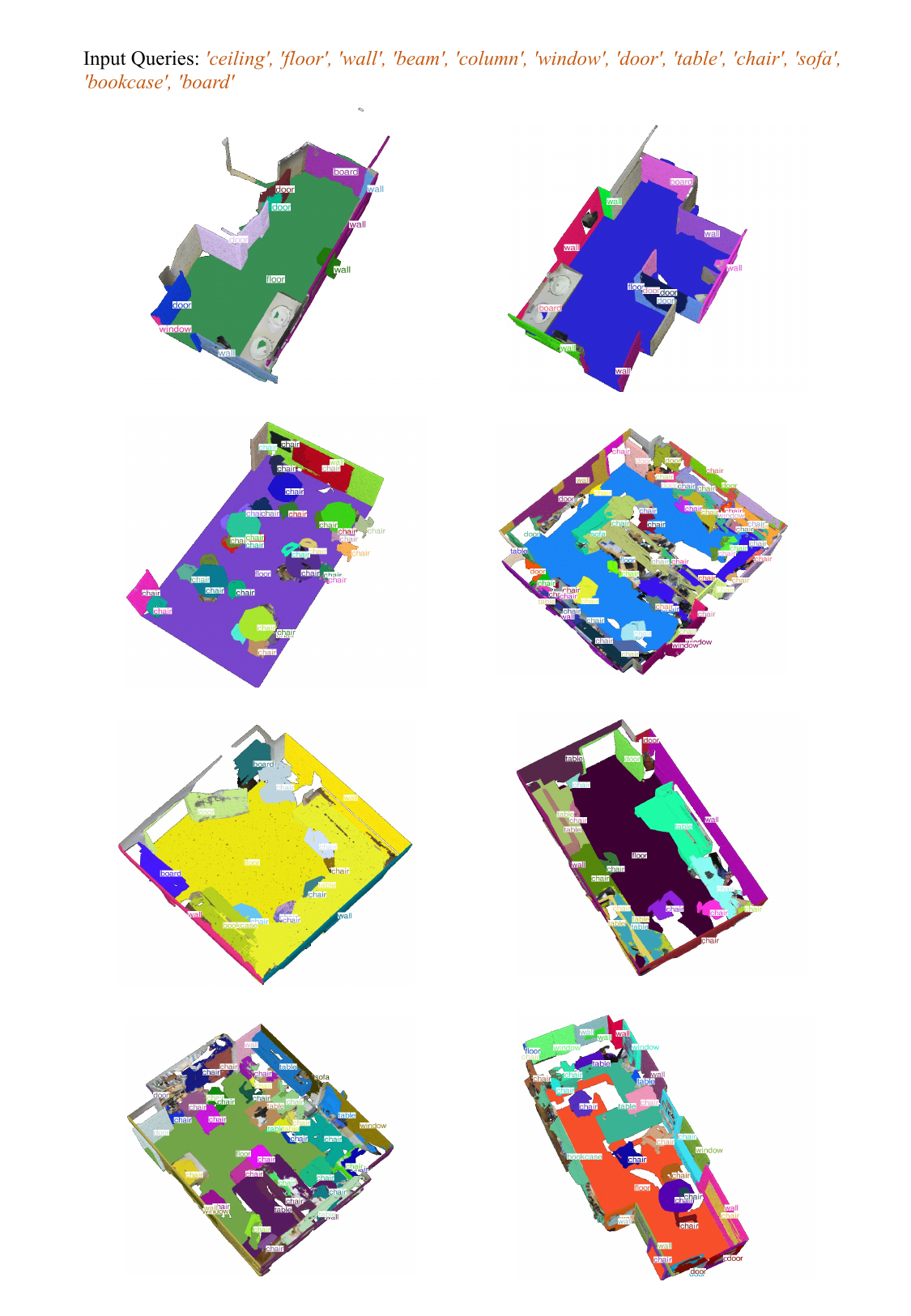}
 \centering
  \caption{\textbf{Open-vocabulary Instance Segmentation Results of S3DIS by OpenIns3D (ODISE).} Instance and class labels are presented in the same color.}
 \label{fig: vis_look_s3dis}
\end{figure*}

\begin{figure*}[t!]
 \centering
 \includegraphics[width=1.0\textwidth]{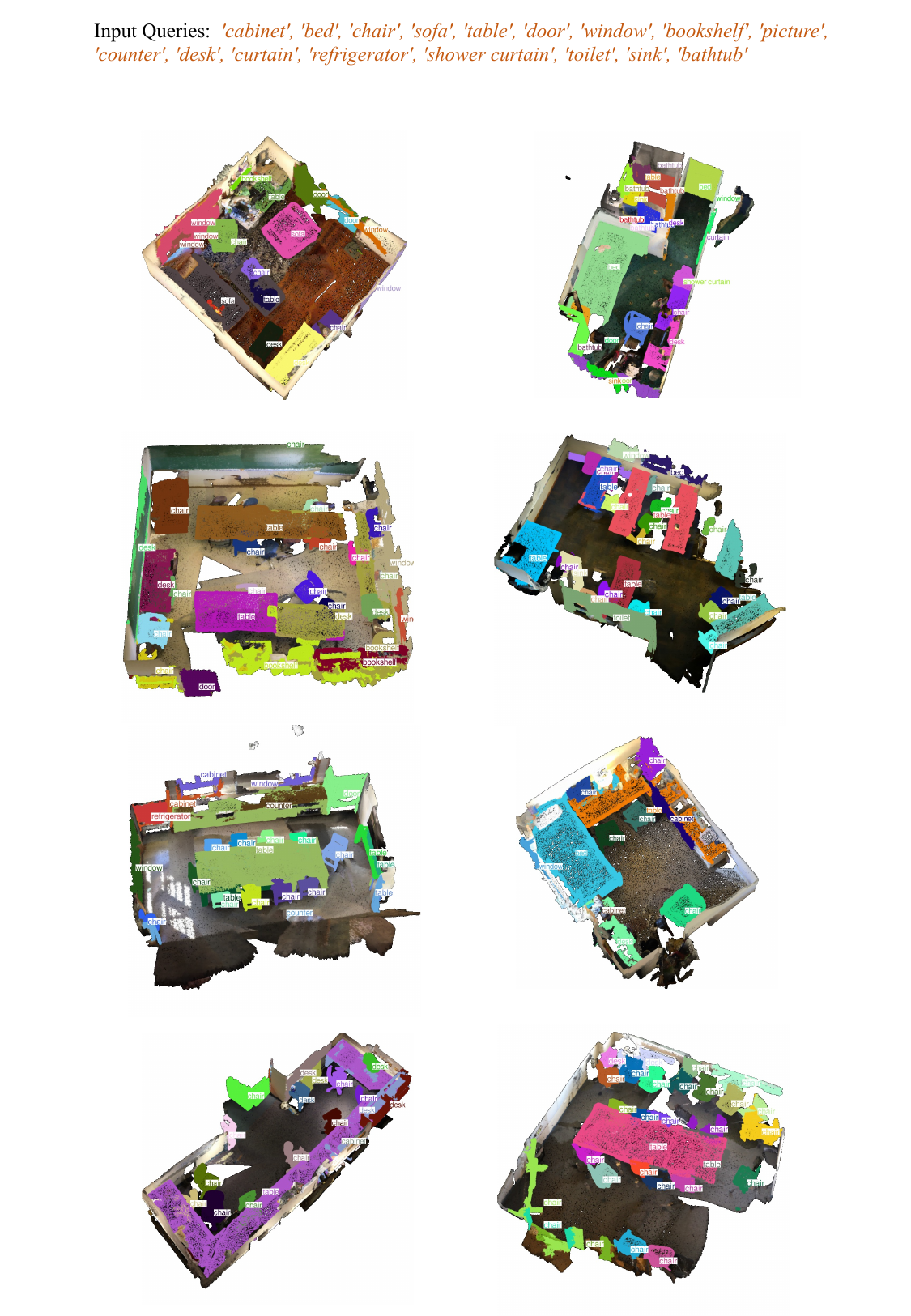}
 \centering
  \caption{\textbf{Open-vocabulary Instance Segmentation Results of ScanNetv2 by OpenIns3D (ODISE).} Instance and class labels are presented in the same color.}
  \label{fig:vis_look_scannet}
\end{figure*}

\begin{figure*}[t!]
 \centering
 \includegraphics[width=1.0\textwidth]{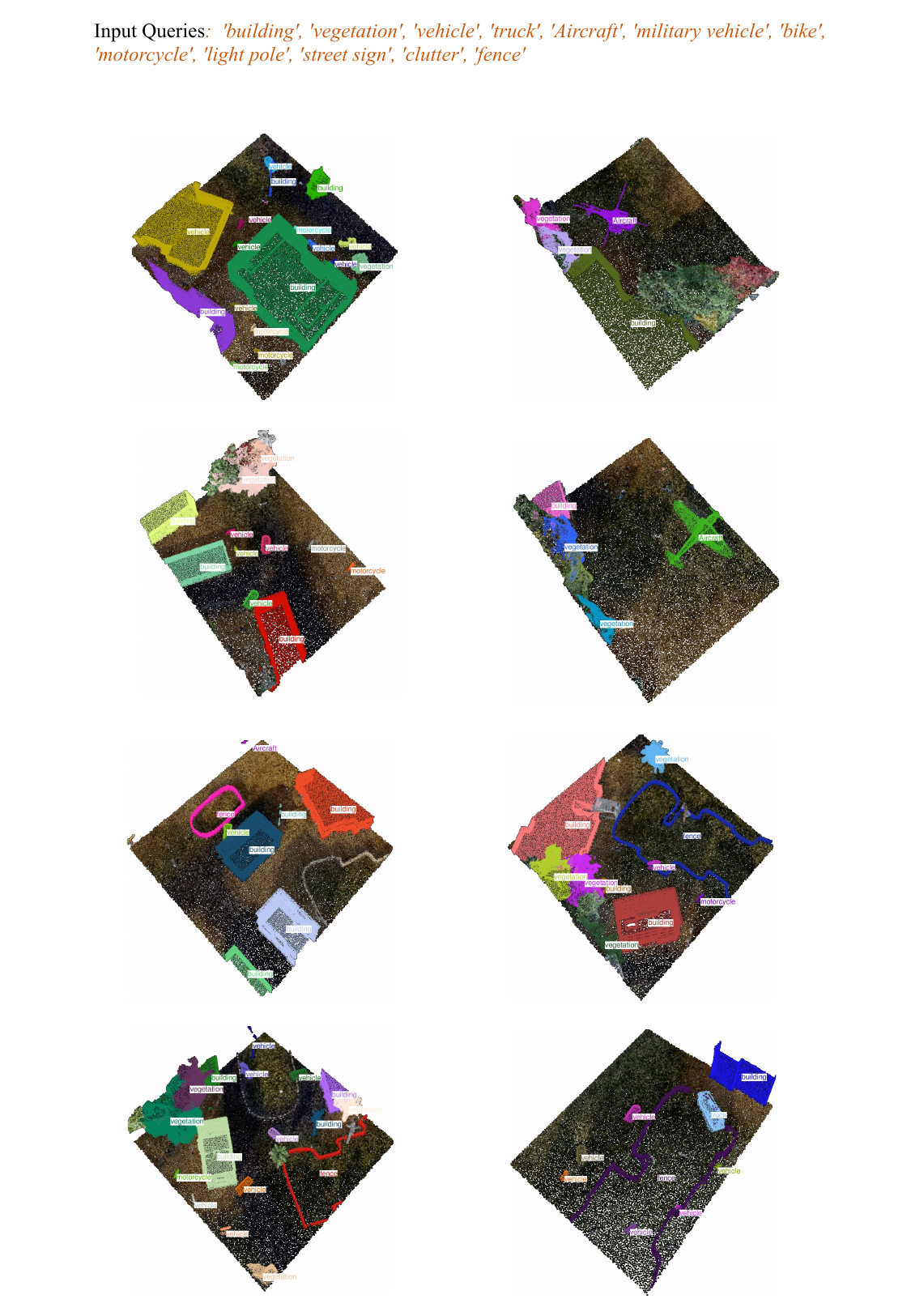}
 \centering
  \caption{\textbf{Open-vocabulary Instance Segmentation Results of STPLS3D by OpenIns3D (ODISE).} Instance and class labels are presented in the same color.}
 \label{fig: vis_look_stpls3d}
\end{figure*}

\noindent\textbf{Lookup results visualization.}
The Lookup module effectively links 2D results with 3D. Here, we present visualizations of its outcomes from all three datasets (Figure \ref{fig: vis_look_s3dis}, \ref{fig:vis_look_scannet}, \ref{fig: vis_look_stpls3d}).

\newpage

\clearpage
\bibliographystyle{splncs04}
\bibliography{main}
\end{document}